\documentclass[11pt]{article}
\usepackage[numbers,sort]{natbib}

\usepackage[margin=1in]{geometry}
\usepackage[T1]{fontenc}
\usepackage[utf8]{inputenc}
\usepackage{lmodern}

\usepackage{graphicx}
\usepackage{booktabs}
\usepackage{amsmath,amssymb}
\usepackage{listings}
\usepackage{authblk}
\usepackage{hyperref}
\usepackage{subcaption}
\usepackage{placeins}
\usepackage{fancyhdr}
\usepackage{xcolor}
\definecolor{linkgreen}{RGB}{0,140,0}

\hypersetup{
    colorlinks=true,
    linkcolor=black,
    citecolor=linkgreen,
    urlcolor=linkgreen
}

\fancypagestyle{firstpage}{
    \fancyhf{}
    \fancyfoot[C]{%
    \scriptsize\setlength{\fboxsep}{1pt}%
    \makebox[\textwidth][c]{%
    \textbf{Code:}~\href{https://github.com/RicharMd/RealMath-Eval}{\fcolorbox{linkgreen}{white}{\strut\nolinkurl{github.com/RicharMd/RealMath-Eval}}}%
    ~|~
    \textbf{Data:}~\href{https://huggingface.co/datasets/RicharMd/RealMath-Eval}{\fcolorbox{linkgreen}{white}{\strut\nolinkurl{huggingface.co/datasets/RicharMd/RealMath-Eval}}}%
    }}

}

\title{RealMath-Eval: Why SOTA Judges Struggle with Real Human Reasoning}

\author[1]{Yiteng Mao}
\author[2]{Kenan Xu}
\author[3]{Yijia Lyu}
\author[4]{Wenhao Li}
\author[5]{Jianlong Chen\thanks{Corresponding author: \texttt{jianlongchen@link.cuhk.edu.cn}}}
\author[2]{Xiangfeng Wang}

\affil[1]{University of Wisconsin--Madison}
\affil[2]{East China Normal University}
\affil[3]{New York University}
\affil[4]{Tongji University}
\affil[5]{The Chinese University of Hong Kong, Shenzhen}
\affil[]{\footnotesize \texttt{mao85@wisc.edu} \quad | \quad \texttt{jianlongchen@link.cuhk.edu.cn}}

\date{}

\begin{document}

\maketitle
\thispagestyle{firstpage}

\begin{center}
\small
\href{https://github.com/RicharMd/RealMath-Eval}{\includegraphics[height=1.2em]{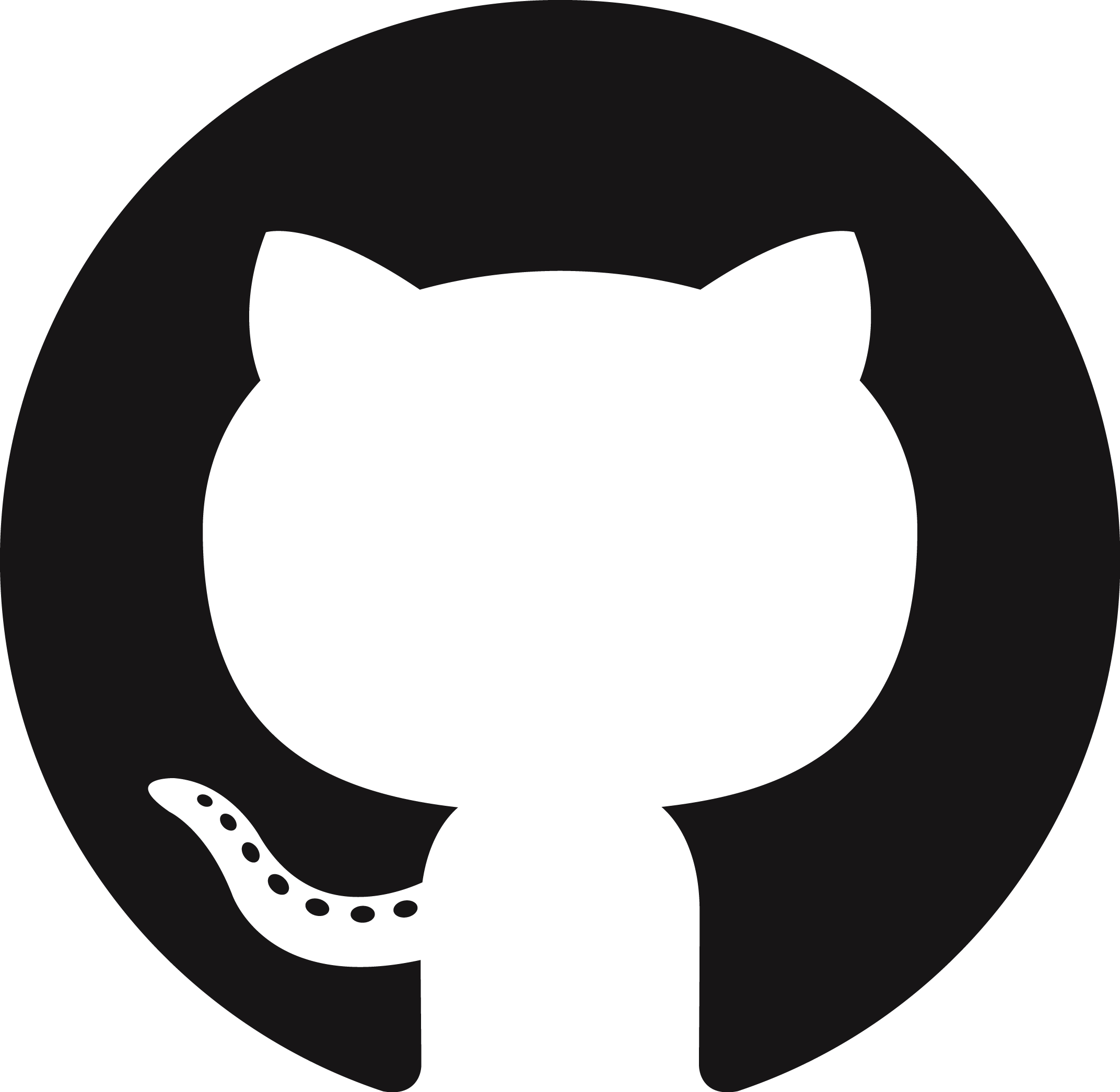}\hspace{0.35em}GitHub Repository}
\hspace{1em}
\href{https://huggingface.co/datasets/RicharMd/RealMath-Eval}{\includegraphics[height=1.2em]{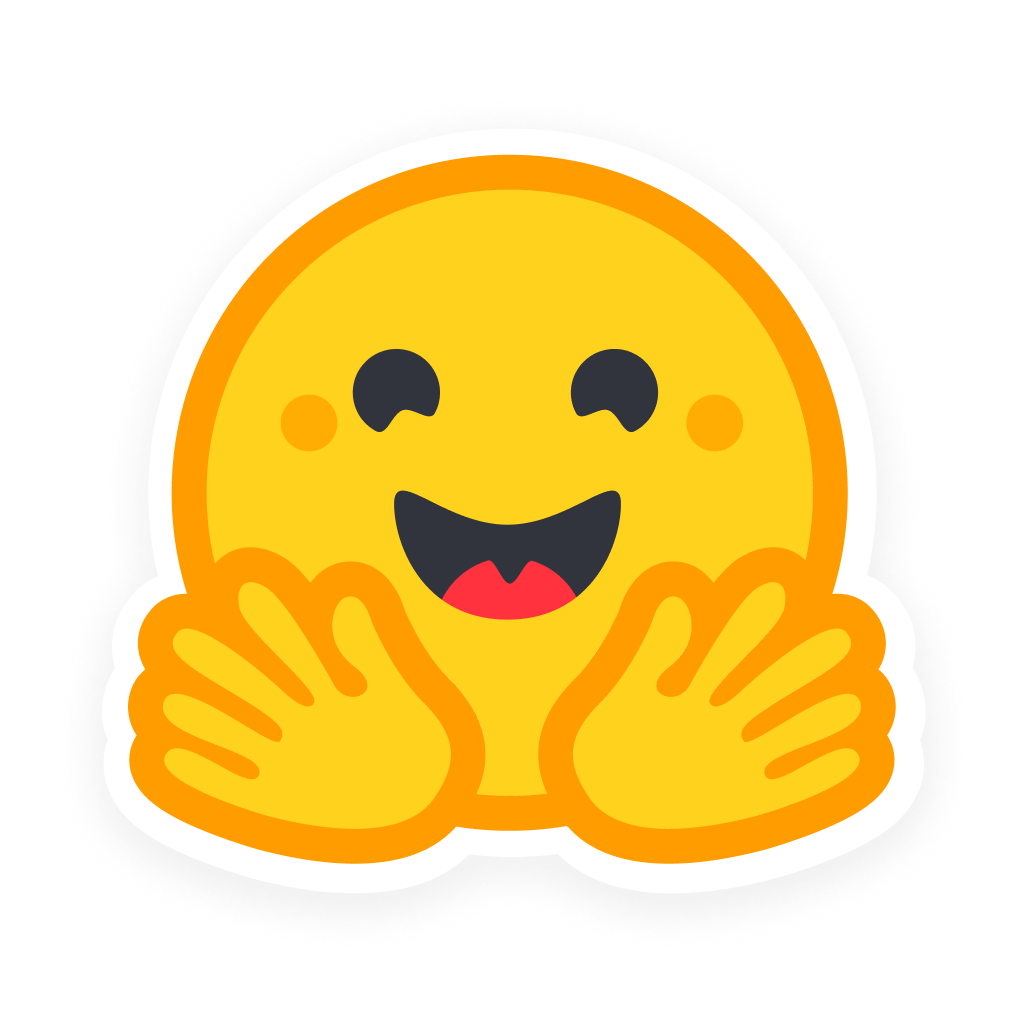}\hspace{0.35em}Hugging Face Dataset}
\end{center}
\vspace{0.5em}

\graphicspath{{exp_images/}}

\begin{abstract}
  While Large Language Models (LLMs) have achieved near-perfect performance in \emph{solving} high-school mathematics, their ability to \emph{evaluate} the diverse reasoning processes of real human students remains under-examined. To bridge this gap, we introduce \textbf{RealMath-Eval}, a rigorously annotated benchmark of 224 real-world exam responses from high schools. Our initial evaluation reveals that even state-of-the-art LLM judges struggle significantly on this task, exhibiting a high Mean Squared Error ($\sim$2.96) against expert human grading. To probe a plausible explanation, we contrast this performance with a control setting where the same judges evaluate synthetic LLM-generated solutions. We identify a stark ``Evaluation Gap'': judges are considerably more accurate and consistent on synthetic text (MSE $\sim$1.17) but struggle to generalize to authentic student reasoning. Through semantic embedding analysis, we find that synthetic errors suffer from a ``structural collapse'' into predictable, low-dimensional linear subspaces, whereas human errors form a more diverse error space. Furthermore, generative probability probes suggest that human reasoning involves significantly higher information-theoretic surprisal, indicating that student reasoning transitions are more out-of-distribution for current models. Finally, we find that surface-level style transfer fails to close this gap. Our findings suggest that current LLM evaluation pipelines relying heavily on synthetic data may not adequately capture the diversity of authentic student mathematical reasoning.
\end{abstract}

\section{Introduction}
The rapid evolution of Large Language Models (LLMs) has led to a paradigm shift in automated evaluation. ``LLM-as-a-Judge'' has become a standard practice, where strong models are deployed to evaluate the quality of weaker models~\citep{Zheng2023JudgingLW,Gu2024ASO}. This approach relies on the implicit assumption that a model capable of \emph{solving} a complex problem is inherently capable of \emph{judging} a solution to that problem, regardless of whether the solution comes from a machine or a human.

\begin{figure*}[!t]
\centering
\includegraphics[width=0.82\textwidth]{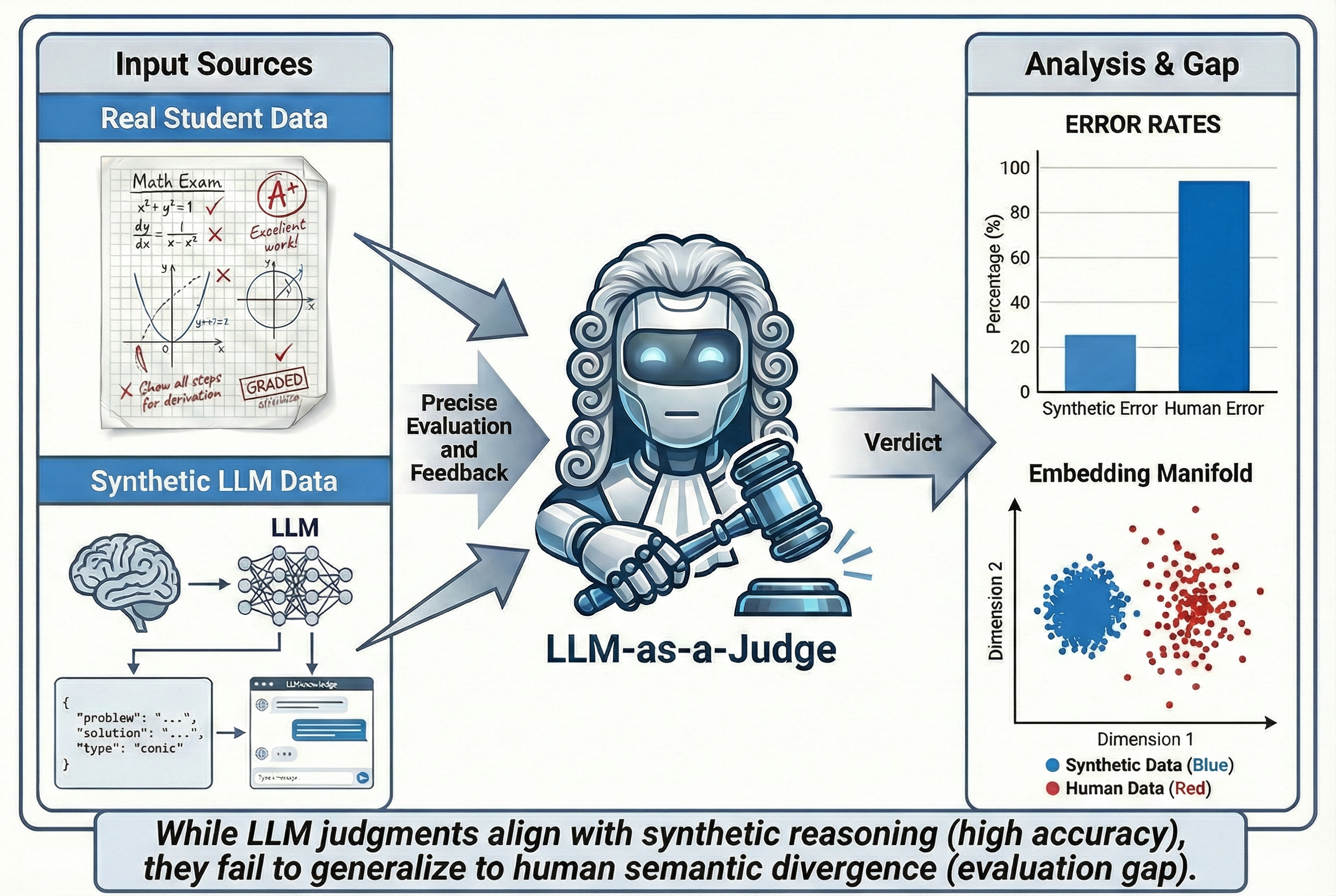}
\caption{Overview of RealMath-Eval and the Evaluation Gap. The benchmark enables a controlled comparison between authentic student reasoning and synthetic LLM solutions, revealing a substantial performance gap that motivates the semantic and LL analyses in the remainder of the paper.}
\label{fig:overview}
\end{figure*}

Current LLM-as-a-Judge research typically falls into two categories: pairwise ranking (comparing two outputs) and pointwise scoring (assigning an absolute quality score)~\citep{Li2024FromGT}. While pairwise ranking is widely used for preference alignment, pointwise scoring is indispensable for applications requiring calibrated judgments, such as reward modeling and rubric-based evaluation.~\citep{Li2024FromGT,tan2024judgebench,malik2025rewardbench}.

However, existing benchmarks for pointwise scoring predominantly focus on synthetic data---using strong models to grade the outputs of weaker models~\citep{tan2024judgebench,malik2025rewardbench}. This creates a closed loop where judges are evaluated on text distributions they are statistically familiar with. While educational NLP has extensively evaluated real student text in Automated Essay Scoring (AES)---ranging from traditional systems relying on shallow linguistic features \citep{ramesh2022automated} to recent LLM-based multi-trait evaluators \citep{lee2024unleashing}---these approaches primarily assess writing quality and rhetorical coherence. In contrast, rubric-grounded evaluation of complex mathematical reasoning requires tracking stepwise deductions, logical validity, and error propagation. 
A critical question thus emerges for alignment: can these judges, optimized for the predictable patterns of machine text, effectively evaluate the \textbf{diverse, noisy, and often idiosyncratic reasoning} of real human students? This capability is essential for reliable reward signals in RLHF and trustworthy AI tutors, yet remains under-examined for authentic student mathematical reasoning.

To address this, we introduce \textbf{RealMath-Eval}, a benchmark constructed from 224 expert-annotated real-world high school exam responses. Our initial experiments reveal a concerning reality: State-of-the-art (SOTA) LLM judges struggle significantly when grading real student solutions, exhibiting a high Mean Squared Error ($\sim$2.96) compared to human ground truth. To probe this failure, we conducted a comparative analysis against a control group of synthetic LLM solutions. This comparison unveiled a stark \textbf{Evaluation Gap}: the same judges are remarkably accurate (MSE $\sim$1.17) when evaluating synthetic text, suggesting that the difficulty arises not merely from mathematical complexity, but more importantly from differences in how humans and models construct erroneous reasoning paths. Figure~\ref{fig:overview} depicts an overview of our analysis, and we summarize our key \textbf{contributions} as follows:
\begin{enumerate}
    \item \textbf{RealMath-Eval Benchmark}: A rigorously annotated dataset from real high-school exams, specifically curated to test evaluation capabilities on diverse human reasoning.
    \item \textbf{Discovery of the Evaluation Gap}: We reveal a stark contrast in judge performance on the same problems: SOTA judges are far more reliable on synthetic LLM reasoning (MSE $\sim$1.17, Failure Rate 12.8\%) than on human reasoning (MSE $\sim$2.96, Failure Rate 28.7\%). This $\sim$2.5x degradation provides empirical evidence for an observed evaluation asymmetry, which we refer to as a potential ``In-group Bias'': current evaluators appear substantially better calibrated to synthetic text distributions than to authentic human reasoning.
    \item \textbf{Mechanism of Failure: The Crystal vs. The Cloud}: To the best of our knowledge, we provide an early analytical account of this gap. 
      Semantic embedding analysis reveals that LLM errors undergo a ``structural collapse'' into stable, low-dimensional ``crystals,'' whereas human errors exhibit high intra-category divergence, forming an unstructured ``cloud.'' Furthermore, generative probes show that human reasoning involves significantly higher information-theoretic surprisal (Logical Likelihood $\approx$ 0.11 vs. 0.33), suggesting a distributional mismatch that current judge pipelines may not fully resolve through surface-level prompting alone.
\end{enumerate}

\section{Related Work}
\label{sec:related_work}

\paragraph{LLM-as-a-Judge and pointwise evaluation.}
LLM-as-a-Judge has become a common paradigm for automated evaluation, especially in pairwise preference comparison and pointwise scoring~\citep{Zheng2023JudgingLW,Gu2024ASO,Li2024FromGT}. Our setting belongs to the pointwise branch, but differs from open-ended quality scoring because the judge must assign an absolute score under a problem-specific mathematical rubric.

\paragraph{Synthetic judge benchmarks.}
Recent judge and reward-model benchmarks often evaluate model outputs against other model outputs~\citep{tan2024judgebench,malik2025rewardbench}. These benchmarks are valuable for controlled evaluation, but they mostly test distributions familiar to contemporary LLMs. RealMath-Eval complements this line of work by comparing the same judges on synthetic LLM solutions and authentic student responses for the same mathematical problems.

\paragraph{Educational NLP and mathematical reasoning assessment.}
Educational NLP has long studied real student text, especially in Automated Essay Scoring and multi-trait writing assessment~\citep{ramesh2022automated,lee2024unleashing}. Mathematical reasoning assessment requires a different form of evaluation: judges must track stepwise deductions, identify missing intermediate claims, apply partial-credit rubrics, and distinguish valid alternative methods from invalid shortcuts. RealMath-Eval focuses on this rubric-grounded mathematical setting rather than general writing quality.

\section{RealMath-Eval Dataset}
\label{sec:benchmark}
To systematically assess the evaluation capability of LLMs on human data, we introduce \textbf{RealMath-Eval}. 

\subsection{Benchmark Construction}
The dataset is built from three batches of high-school assessments (2025-09-28, 2025-10-17, 2025-10-24), covering advanced topics such as \textbf{trigonometric functions, sequences, derivatives, and conic sections}. We process the handwritten responses with an OCR-and-correction pipeline, remove unusable artifacts, and retain a pool of 359 viable candidates. From this pool, we uniformly sample 16 representative responses for each of 14 problems, yielding a final benchmark of 224 samples. Detailed construction steps are deferred to Appendix~\ref{sec:appendix:pipeline}. Each benchmark sample contains the original problem statement, the student's step-by-step solution, and the reference answer. Every sample is annotated with an expert-assigned \textbf{Ground Truth Score} and \textbf{step-wise scoring labels}, enabling both overall evaluation and fine-grained error localization. The benchmark also spans a broad range of performance levels, with 59 low-performing samples (26.34\%), 97 medium-performing samples (43.30\%), and 68 high-performing samples (30.36\%). The full data release format is provided in Appendix~\ref{sec:appendix:data_release}.

\subsection{Control Group: Synthetic LLM Solutions}
To establish a controlled comparison, we construct a size-matched synthetic dataset of 224 LLM-generated solutions, one for each authentic student sample. Synthetic responses are generated by a diverse set of low-, medium-, and high-tier models at varied temperatures to mimic a range of student ability levels; the full model list and temperature settings are reported in Appendix~\ref{sec:appendix:synthetic_details}. All synthetic solutions are produced under the same ``Student Persona'' prompt and then annotated by the same expert human process used for the real student data, ensuring a fair comparison.

\section{Methodology}
\label{sec:methodology}

We propose a unified framework to evaluate and analyze the capability of LLMs in judging mathematical reasoning. Our methodology consists of a formal task definition, a standardized judge configuration, and a suite of analytical probes designed to dissect the ``Evaluation Gap.''

\subsection{Task Formulation}
We formulate the evaluation task as a pointwise scoring problem~\citep{Li2024FromGT}. Let $P$ be a mathematical problem, $R$ be the reference solution with a step-wise rubric, and $S$ be a student's response. The judge model $\mathcal{M}$ takes the tuple $(P, R, S)$ as input and outputs a scalar score $\hat{y}$ and a reasoning rationale $E$:
\begin{equation}
    (\hat{y}, E) = \mathcal{M}(P, R, S)
\end{equation}
The goal is to minimize the discrepancy between the predicted score $\hat{y}$ and the ground truth score $y_{gt}$ assigned by human experts. We quantify this discrepancy using Mean Squared Error (MSE) and a fine-grained \textbf{Failure Rate} ($\text{FR}_\delta$), defined as the percentage of cases where the absolute score deviation $|\hat{y} - y_{gt}| \ge \delta$. By evaluating across multiple thresholds ($\delta \in \{1, 2, 3, 4, 5\}$), we capture a comprehensive spectrum of evaluation errors, ranging from minor grading variances to catastrophic misjudgments. We use $\text{FR}_2$ as the main severe-error indicator because a two-point deviation in these step-wise rubrics typically corresponds to over-awarding or missing a substantive reasoning milestone, while still reporting $\text{FR}_1$--$\text{FR}_5$ for sensitivity.

\subsection{Judge Configuration and Prompting}
We employ state-of-the-art LLMs (Gemini 3 Pro Preview, GPT-5.2, Qwen 3.5 Plus, and DeepSeek-V3.2) as judges~\citep{gemini3pro2025,gpt52_2025,qwen3.5,liu2025deepseek}. Using MASLab~\citep{ye2025maslab}, we implement a \textbf{Chain-of-Thought (CoT)}~\citep{wei2022chain} ``Reason-then-Score'' prompting strategy (Appendix~\ref{sec:appendix:prompts}) in which the model first analyzes the student's steps against the rubric, then identifies correct and incorrect reasoning, and finally assigns a cumulative score.

\subsection{Meta-Evaluation Protocol}
To qualitatively understand \emph{why} the judge fails, we apply a two-stage attribution workflow on samples with significant disagreement ($\Delta \ge 2$): a prompt-based meta-evaluation classifier produces initial labels, which are then corrected through expert human-in-the-loop review (Appendix~\ref{sec:appendix:prompts}, Appendix~\ref{sec:appendix:hitl}). The corrected cases are grouped into four categories: \textbf{Error Severity}, \textbf{Process Norms}, \textbf{Logical Rigor}, and \textbf{Insight Recognition}. Detailed category definitions are provided in Appendix~\ref{sec:appendix:meta_categories}. This taxonomy allows us to move beyond simple error rates and diagnose recurring rubric-level mismatches between judges and human graders.

\subsection{Analytical Framework}
To probe the possible sources of evaluation failures, we employ two complementary probes on a focused subset of data. We first extract coarse-grained error segments from the benchmark's step-wise annotations (Human $n=278$, Synthetic $n=328$), and then split these segments into atomic reasoning steps for the micro-level analysis. Additional preprocessing details are deferred to Appendix~\ref{sec:appendix:analysis_details}.

\textbf{(1) Semantic Structural Probe (Macro-level).} We map each error segment $s_i \in \mathcal{S}_{err}$ to an embedding vector $\mathbf{h}_i = E(s_i) \in \mathbb{R}^d$ using \textbf{Qwen3-Embedding-8B}~\citep{qwen3-embedding}. To capture both local and global structure, we examine these embeddings along four complementary dimensions: \textbf{Local Dispersion} (mean $k$-nearest-neighbor distance~\citep{loftsgaarden1965nonparametric}), \textbf{Cluster Separation} (HDBSCAN / GMM \& Silhouette Score~\citep{mcinnes2017hdbscan,campello2013density}), \textbf{Global Spatial Pattern} (pairwise-distance heatmaps), and \textbf{Subspace Dimensionality} (PCA explained variance). The detailed definitions of these statistics are provided in Appendix~\ref{sec:appendix:analysis_details}.

\textbf{(2) Generative Predictability Probe (Micro-level).} A key component of our approach is the computation of predictability at the micro-level, which captures how ``surprising'' a specific human reasoning step is to a causal language model~\citep{meister2020if}.

\textbf{Step-wise Transition Probabilities.} We leverage the internal generation signals of a causal language model to evaluate the predictability of reasoning paths. Given an error segment parsed into an ordered sequence of atomic steps $S = (s_0, s_1, \dots, s_{N-1})$, we define the context prefix at step $k$ as $C_k = \bigoplus_{i=0}^k s_i$. We feed $C_k$ into a causal probe model $\mathcal{M}$ (\textbf{Qwen3-8B}~\citep{qwen3technicalreport}, chosen for its strong math reasoning capabilities and open access to internal logits)
 
  to extract the next-token logits. To map these internal representations into a normalized probability space, we apply a softmax function over the vocabulary $\mathcal{V}$:
\begin{equation}
    p(v \mid C_k) = \text{softmax}(\text{Logits}(C_k))[v], \quad \forall v \in \mathcal{V}
\end{equation}
This probability quantifies how likely the model is to generate token $v$ as the immediate continuation of the context $C_k$.

\textbf{Logical Likelihood (LL) Score.} To formally quantify this step-level divergence, we introduce the Logical Likelihood (LL) score as a metric tailored for our evaluation framework. Based on the probability distribution, we evaluate the model's anticipation of the actual next step $s_{k+1}$ produced by the student. Let $\mathcal{T}_{k+1}$ be the set of valid token IDs composing $s_{k+1}$. We define the transition Logical Likelihood as the maximum probability assigned to any token within the actual next step:
\begin{equation}
    \text{LL}(s_k \to s_{k+1}) = \max_{t \in \mathcal{T}_{k+1}} p(t \mid C_k)
\end{equation}
A lower LL score indicates that even the most likely token in the student's actual reasoning step was assigned a low probability by the model (i.e., a high ``surprise'' factor), reflecting idiosyncratic, ``jumpy'' human logic. We aggregate this metric for each segment by taking the maximum transition LL: $\max_k \text{LL}(s_k \to s_{k+1})$.

We use the maximum token probability as a task-specific transition probe rather than as a replacement for sequence likelihood. Mathematical derivations often contain many formatting and connective tokens, while the predictability of a reasoning transition may hinge on a small number of decisive mathematical tokens. The max formulation therefore captures whether the probe model assigns high probability to any part of the next reasoning step. As a robustness check, we also compute step-level perplexity, a standard language-model evaluation measure~\citep{chen-goodman-1996-empirical}, in Appendix~\ref{sec:appendix:ll_ppl}, which yields the same directional conclusion.

\section{Experiments}
We first evaluate SOTA judges on real student data, then use a controlled comparison to examine whether the difficulty concentrates on authentic human reasoning.

\subsection{Experimental Setup}
We evaluated four SOTA judge models: Gemini 3 Pro Preview, GPT-5.2, Qwen 3.5 Plus, and DeepSeek-V3.2 \citep{gemini3pro2025,gpt52_2025,qwen3.5,liu2025deepseek}. All models were prompted with Chain-of-Thought (CoT) \citep{wei2022chain} to ensure maximum reasoning capability. The evaluation task involved pointwise scoring based on each problem's rubric-defined full score. We use two datasets: the RealMath-Eval benchmark with 224 authentic student responses, and a control set of 224 synthetic samples generated by diverse LLMs (Section~\ref{sec:exp:gap}).

\subsection{Evaluation Metrics}
\label{sec:exp:metrics}
For the primary evaluation task, we report \textbf{Mean Squared Error (MSE)}, \textbf{Exact Match}, and a family of \textbf{Failure Rates} ($\text{FR}_\delta$), where $\text{FR}_\delta$ denotes the proportion of cases with score deviation at least $\delta$ and $\delta \in \{1,2,3,4,5\}$. Analysis-specific metrics such as Silhouette Score and Logical Likelihood (LL) are defined in Section~\ref{sec:methodology}.

\subsection{Main Results on RealMath-Eval}
\label{sec:exp:results}

We first benchmark the judges on \textbf{RealMath-Eval}. Table~\ref{tab:main_results} reports MSE, Exact Match, and failure rates across thresholds, revealing substantial vulnerability across all tested LLMs.

\begin{table}[ht]
    \centering
    \caption{Main results on RealMath-Eval. All judges exhibit a substantial long tail of severe misjudgments ($\text{FR}_2 \ge 26\%$).}
    \label{tab:main_results}
    {\setlength{\tabcolsep}{4pt}\renewcommand{\arraystretch}{0.95}
    \resizebox{\linewidth}{!}{
    \begin{tabular}{lccccccc}
    \toprule
    \textbf{Judge Model} & \textbf{MSE ($\downarrow$)} & \textbf{Exact Match ($\uparrow$)} & \textbf{$\text{FR}_1$ ($\downarrow$)} & \textbf{$\text{FR}_2$ ($\downarrow$)} & \textbf{$\text{FR}_3$ ($\downarrow$)} & \textbf{$\text{FR}_4$ ($\downarrow$)} & \textbf{$\text{FR}_5$ ($\downarrow$)} \\
    \midrule
    \textbf{Qwen 3.5 Plus} & \textbf{2.67} & \textbf{37.9\%} & \textbf{62.1\%} & \textbf{26.3\%} & \textbf{11.2\%} & \textbf{4.0\%} & \textbf{1.8\%} \\
    \textbf{Gemini 3 Pro} & 2.96 & 37.7\% & 62.3\% & 28.7\% & 12.6\% & 4.9\% & 2.2\% \\
    \textbf{GPT-5.2}       & 5.28 & 33.9\% & 66.1\% & 41.1\% & 21.4\% & 11.2\% & 7.6\% \\
    \textbf{DeepSeek-V3.2} & 7.57 & 25.9\% & 74.1\% & 48.7\% & 30.8\% & 20.5\% & 11.2\% \\
    \bottomrule
    \end{tabular}
    }
    }
    \end{table}

\textbf{Human Inter-Rater Ceiling.}
To contextualize these scores, we conducted a blind inter-rater consistency check on a random subset of 50 authentic responses. A second evaluator re-annotated the samples using the same rubric as the original expert annotations. The human-human agreement reached 82.0\% Exact Match with an MSE of 0.18, and no severe disagreements under $\text{FR}_2$ (0\%). This ceiling suggests that the low Exact Match of LLM judges is not merely a consequence of inherent rubric ambiguity; rather, the models exhibit substantially larger deviations than human graders applying the same scoring standard.

\textbf{Universal Struggle with Human Data.} The most prominent finding is that judge performance is universally far from ideal. Across all models, the Exact Match rate never exceeds 38\%, indicating a pervasive difficulty in perfectly aligning with expert grading. More critically, the Exact Match metric alone does not capture the severity of the misjudgments. When evaluating the Failure Rate ($\Delta \ge 2$), we observe that even the strongest models significantly misjudge over a quarter of the student responses (26.3\% for Qwen 3.5 Plus and 28.7\% for Gemini 3 Pro). This high rate of severe deviation suggests that the models are not merely making minor point-deduction errors, but rather struggling to consistently evaluate the logical structure of the students' reasoning.

\textbf{High Variance Across SOTA Models.} We also observe substantial variance in robustness among the leading models. While Qwen 3.5 Plus and Gemini 3 Pro maintain moderate MSE scores (2.67 and 2.96, respectively), other models struggle far more severely. Notably, GPT-5.2 exhibits a surprisingly high Failure Rate of 41.1\%, and DeepSeek-V3.2 fails on nearly half of the human samples (48.7\%, MSE 7.57). This performance gap suggests that general capability scaling (e.g., GPT-5.2) or strong coding/math reasoning capabilities (e.g., DeepSeek-V3.2) do not automatically translate to the robust evaluation of idiosyncratic human logical errors.

\textbf{Failure Modes in High-Disagreement Cases.} To better understand the nature of these failures, we performed a meta-evaluation on the 64 Gemini 3 Pro cases with significant disagreement ($\Delta \ge 2$), using the attribution categories introduced in Section~\ref{sec:methodology} (full definitions in Appendix~\ref{sec:appendix:meta_categories}). The most common failure mode was \textbf{Category A (Error Severity)} with 26 cases (40.6\%), where the judge often assigns unwarranted ``partial credit'' for non-existent logic in calculation errors (follow-through), whereas humans penalize strictly according to the rubric. \textbf{Category C (Logical Rigor)} followed with 20 cases (31.3\%), where humans penalize logical gaps (e.g., missing sufficient conditions) more severely, while the judge is overly lenient; \textbf{Category B (Process Norms)} accounted for 18 cases (28.1\%), involving non-standard formatting or skipped steps that confused the judge. This pattern suggests that the judge favors a ``roughly correct'' heuristic over the stricter notions of severity and rigor used in human mathematical grading. Overall, these failure patterns further highlight that evaluating authentic student reasoning remains an unsolved challenge for current SOTA models. We next ask whether this difficulty is specific to human reasoning by comparing the same judges against synthetic solutions.

\subsection{The Evaluation Gap: Human vs. Synthetic Reasoning}
\label{sec:exp:gap}

The poor performance on RealMath-Eval raises a critical question: \textbf{\emph{Are these models simply bad at judging math, or is there something specific about human reasoning that confuses them?}}

To answer this, we evaluated the same judges on our \textbf{Control Dataset} of synthetic LLM-generated solutions. Table~\ref{tab:gap_comparison} compares MSE, Exact Match, and $\text{FR}_2$ across the two domains, allowing us to clearly examine the ``Evaluation Gap.''

\begin{table}[ht]
    \centering
    \caption{The Evaluation Gap. All judges perform better on synthetic LLM solutions than on authentic student reasoning.}
    \label{tab:gap_comparison}
    {\footnotesize
    \setlength{\tabcolsep}{4pt}
    \renewcommand{\arraystretch}{0.95}
    \begin{tabular}{llcc}
    \toprule
    \textbf{Judge Model} & \textbf{Metric} & \textbf{Synthetic (LLM)} & \textbf{Human (Student)} \\
    \midrule
    \textbf{Gemini 3 Pro} 
    & MSE ($\downarrow$) & \textbf{1.24} & 2.96 \\
    & Exact Match ($\uparrow$) & \textbf{67.6\%} & 37.7\% \\
    & $\text{FR}_2$ ($\downarrow$) & \textbf{12.8\%} & 28.7\% \\
    \midrule
    \textbf{GPT-5.2} 
    & MSE ($\downarrow$) & \textbf{2.55} & 5.28 \\
    & Exact Match ($\uparrow$) & \textbf{51.6\%} & 33.9\% \\
    & $\text{FR}_2$ ($\downarrow$) & \textbf{26.9\%} & 41.1\% \\
    \midrule
    \textbf{Qwen 3.5 Plus} 
    & MSE ($\downarrow$) & \textbf{1.42} & 2.67 \\
    & Exact Match ($\uparrow$) & \textbf{57.8\%} & 37.9\% \\
    & $\text{FR}_2$ ($\downarrow$) & \textbf{17.4\%} & 26.3\% \\
    \midrule
    \textbf{DeepSeek-V3.2} 
    & MSE ($\downarrow$) & \textbf{7.12} & 7.57 \\
    & Exact Match ($\uparrow$) & \textbf{37.4\%} & 25.9\% \\
    & $\text{FR}_2$ ($\downarrow$) & \textbf{43.4\%} & 48.7\% \\
    \bottomrule
    \end{tabular}
    }
    \end{table}

Table~\ref{tab:gap_comparison} reveals a consistent cross-model contrast: all judges perform better on synthetic data than on human data, with Qwen 3.5 Plus and Gemini 3 Pro showing the strongest overall performance while still retaining a sizable Human-vs-Synthetic gap.

This cross-domain degradation provides empirical support for an evaluation asymmetry:

LLM judges are more effective on synthetic reasoning but struggle to bridge the gap to the diverse reasoning of real students. To better understand this divergence, we next analyze it through macro- and micro-level probes.

\section{Analysis: Convergence vs. Divergence}
\label{sec:analysis}
Having established that the gap is concentrated on authentic human reasoning, we now turn to our analytical probes to visualize the \emph{geometry} of this divergence. We hypothesize that LLM errors are ``convergent'' (collapsing into known statistical patterns), while human errors are ``divergent'' (high-variance and idiosyncratic).

\subsection{Semantic Embedding Analysis}
To map the geometric ``cluster of mistakes,'' we embedded error segments using Qwen3-Embedding-8B~\citep{qwen3-embedding}. Table~\ref{tab:clustering_stats} summarizes the resulting space through local dispersion, cluster separation, and low-dimensional structure.

To check that the observed geometric gap is not an artifact of a single embedder, we repeat the embedding analysis with a Gemma-family KaLM embedder~\citep{hu2025kalmembedding} and a No-Qwen synthetic subset that removes Qwen-family generators from the synthetic side. The same directional Human-vs-Synthetic contrast is preserved; details are reported in Appendix~\ref{sec:appendix:probe_robustness}.

\begin{table}[ht]
    \centering
\caption{Embedding space statistics. Synthetic LLM errors are more clustered and lower-dimensional than human errors.}
\label{tab:clustering_stats}
\footnotesize
\setlength{\tabcolsep}{5pt}
\renewcommand{\arraystretch}{0.95}
\begin{tabular}{lcc}
\toprule
\textbf{Metric} & \textbf{Human Errors} & \textbf{LLM Errors} \\
\midrule
Avg NN Dist ($\downarrow$) & 0.1761 & \textbf{0.1144} \\
Silhouette Score ($\uparrow$) & 0.3339 & \textbf{0.5180} \\
PCA Top-10 Variance ($\uparrow$) & 56.15\% & \textbf{62.46\%} \\
Optimal GMM K & 6 & 27 \\
\bottomrule
\end{tabular}
\end{table}

\begin{figure}[htbp]
    \centering
    \includegraphics[width=0.8\linewidth]{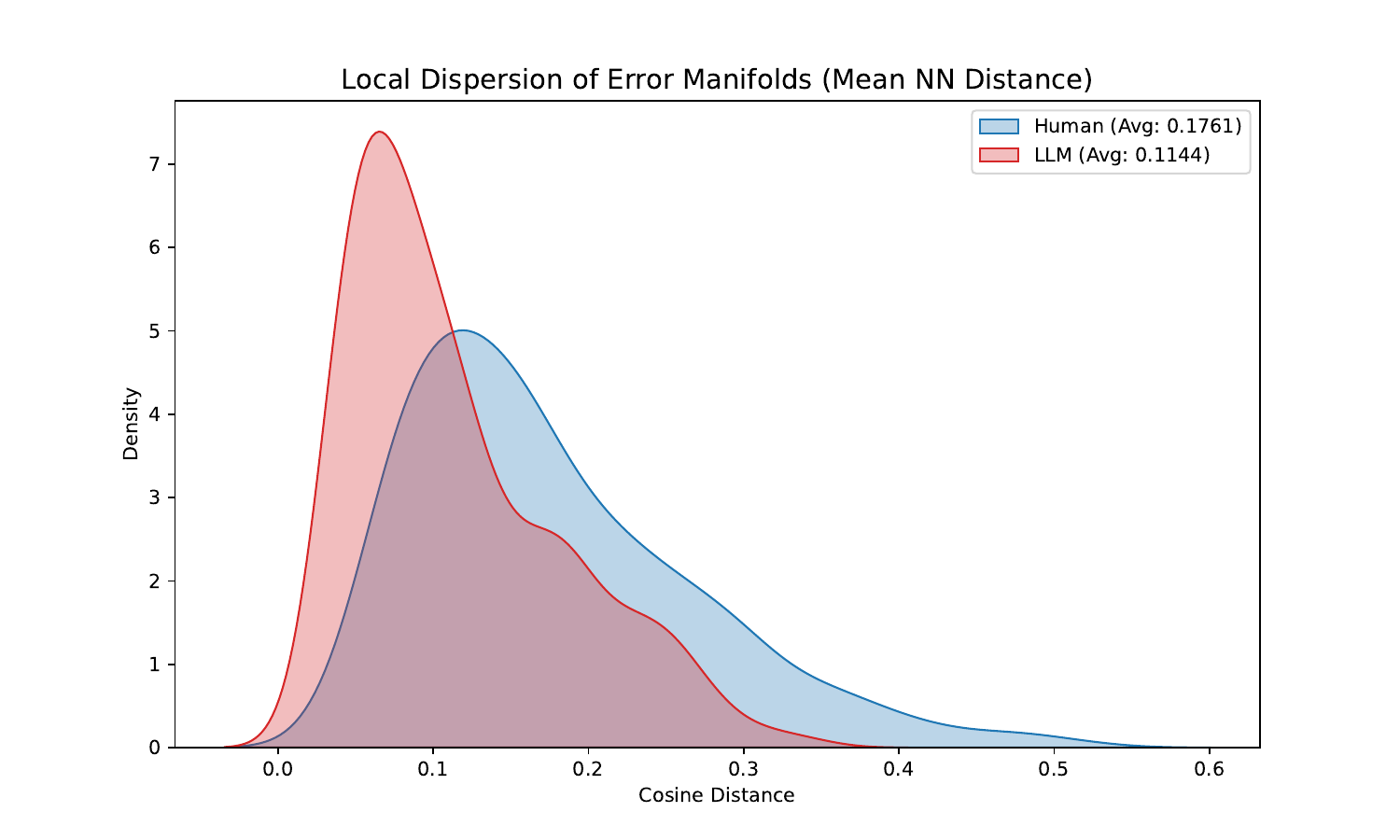}
    \caption{\textbf{Local dispersion of error distributions.} Synthetic LLM errors form a sharper, lower-variance peak than authentic human errors (mean $0.114$ vs.\ $0.176$).}
    \label{fig:local_dispersion}
\end{figure}

\textbf{The ``Crystal'' vs. The ``Cloud'' Metaphor.} As shown in Table~\ref{tab:clustering_stats}, LLM errors exhibit significantly higher \textbf{Cluster Separation} (Silhouette $0.518$ vs. $0.334$) and reduced \textbf{Local Dispersion} (Avg NN Dist $0.1144$ vs. $0.1761$). We use ``Crystal'' and ``Cloud'' as an illustrative metaphor for this observed geometric contrast: when an LLM errs, its errors tend to concentrate into dense, predictable, low-variance semantic patterns, whereas authentic human errors are more diffuse, higher-variance, and less rigidly clustered. This contrast poses a challenge for judge models calibrated primarily on machine-generated distributions.

\begin{figure*}[t]
    \centering
    \begin{tabular}{@{}c@{\hspace{0.02\textwidth}}c@{\hspace{0.02\textwidth}}c@{}}
        \includegraphics[width=0.32\textwidth]{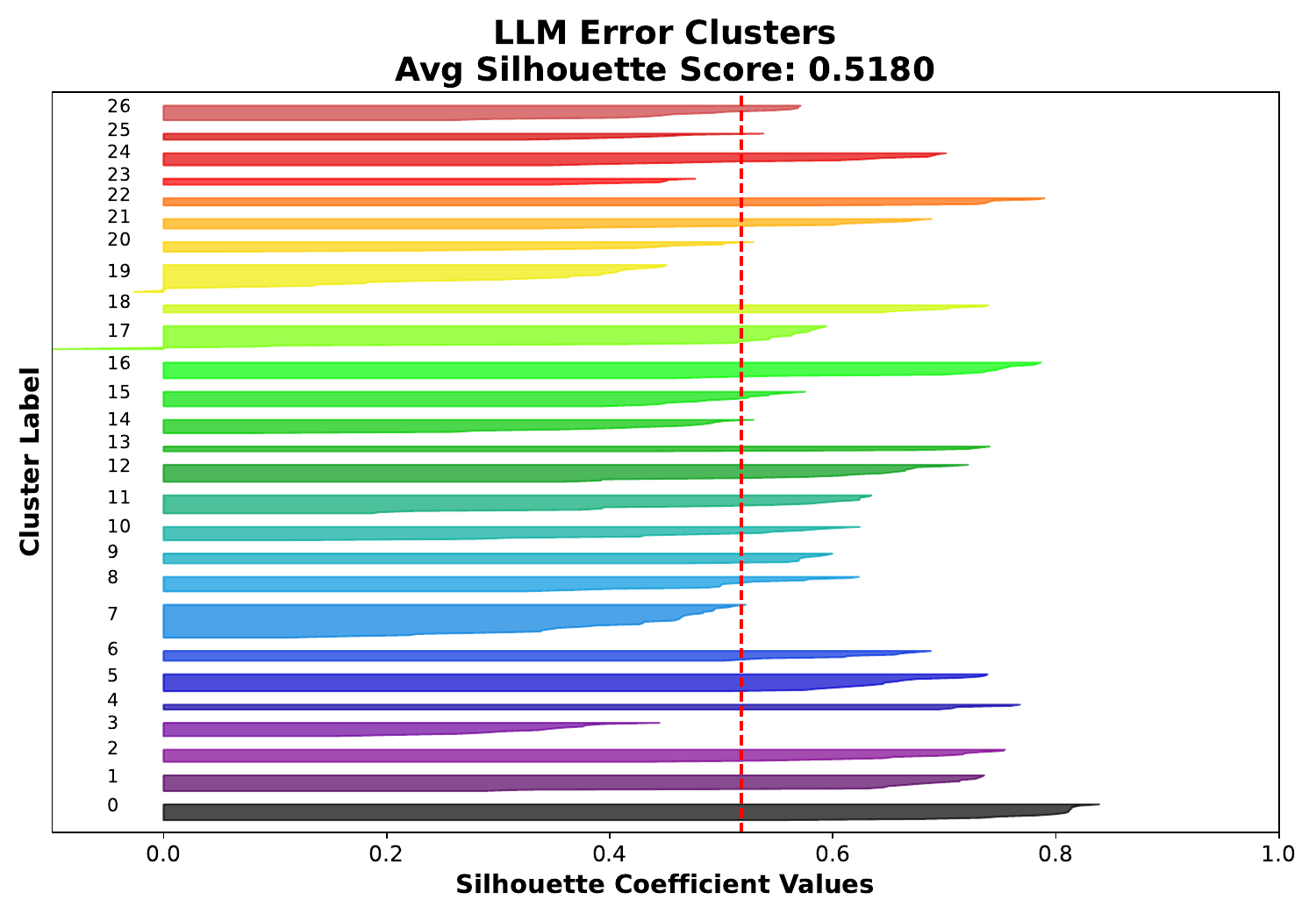} &
        \includegraphics[width=0.32\textwidth]{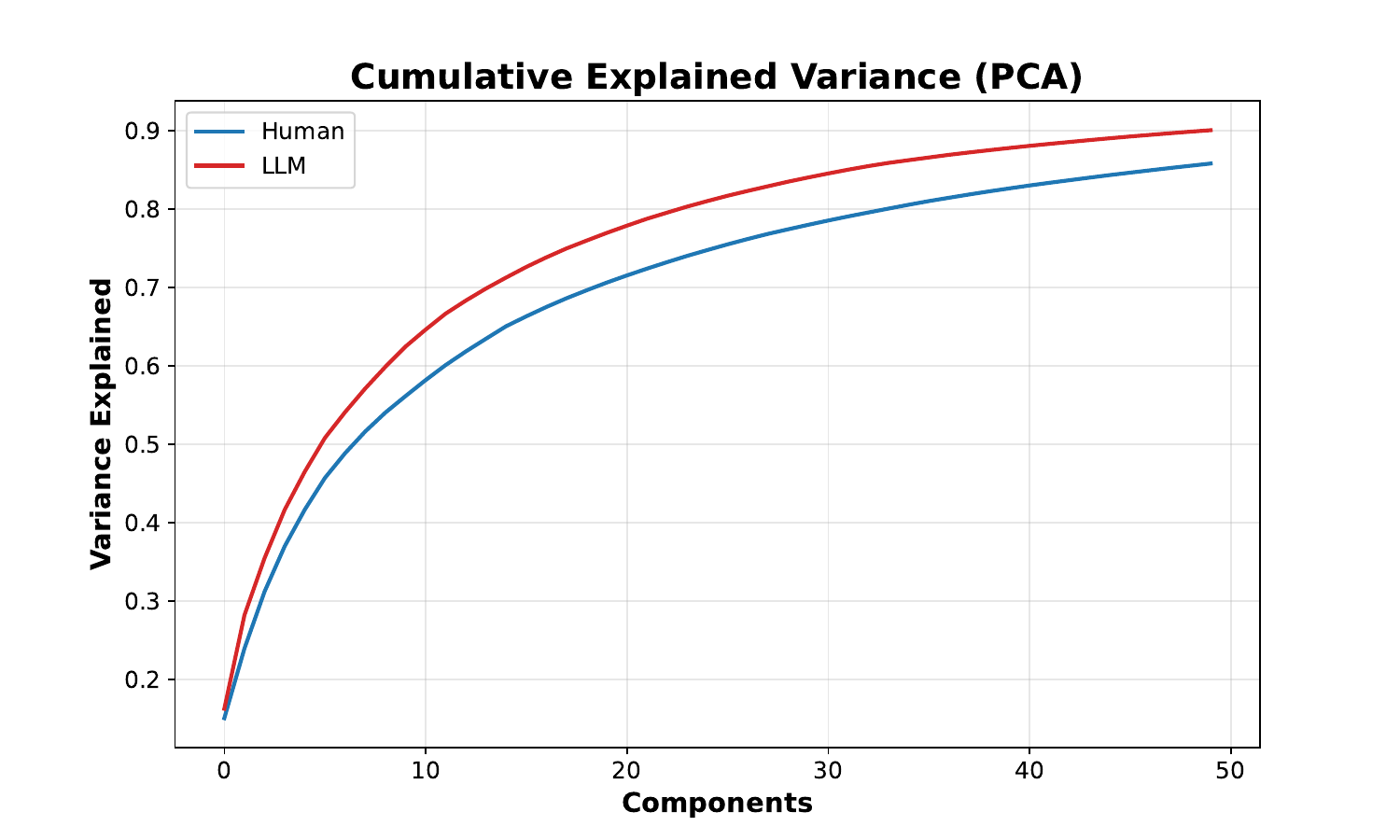} &
        \includegraphics[width=0.32\textwidth]{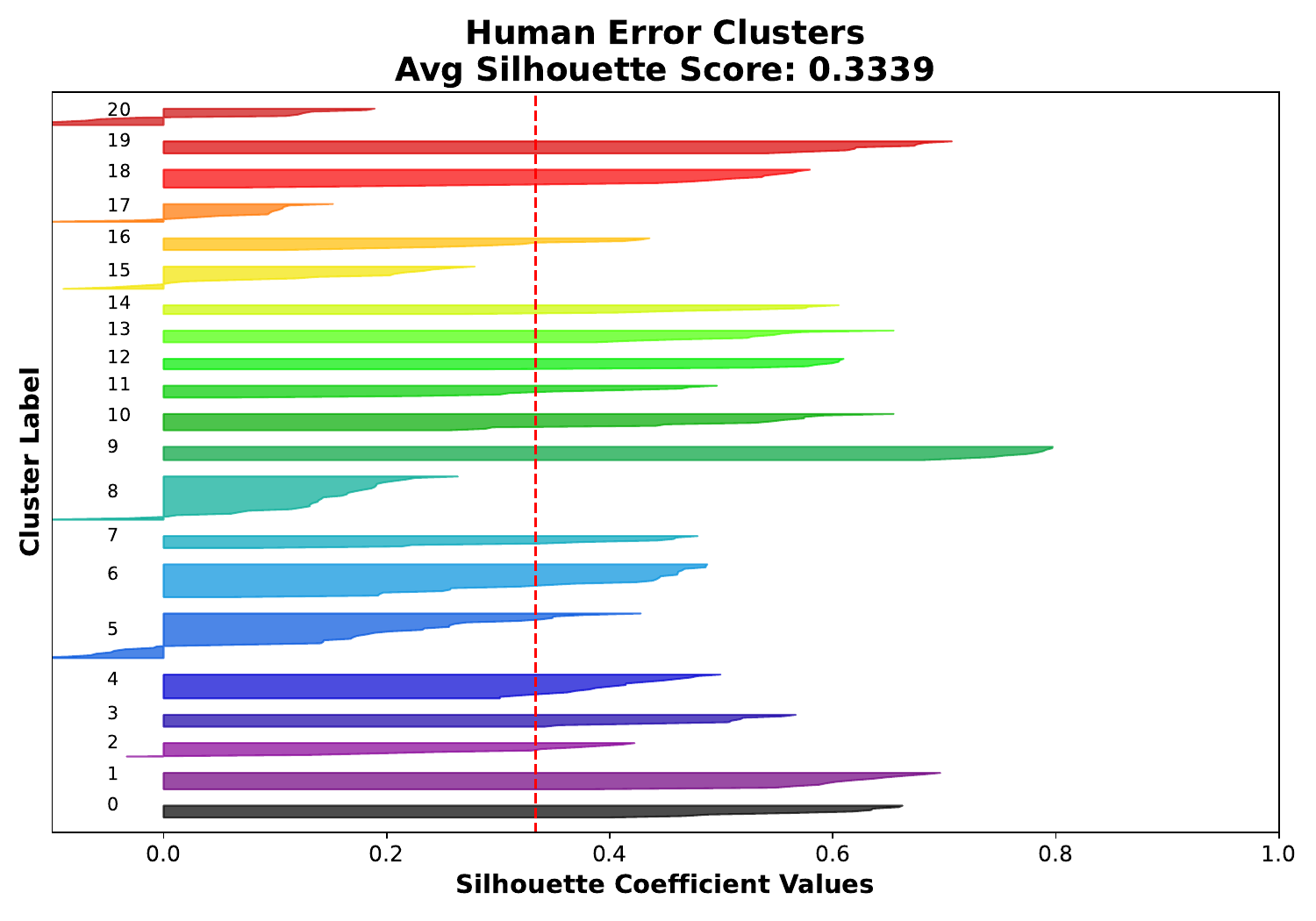} \\[4pt]
        \multicolumn{1}{c}{\small (a) Synthetic (LLM) Silhouette} &
        \multicolumn{1}{c}{\small (b) PCA Embedding Space} &
        \multicolumn{1}{c}{\small (c) Authentic (Human) Silhouette}
    \end{tabular}
    \caption{\textbf{Semantic structure of errors.} Synthetic errors form tighter, more coherent clusters, whereas human errors remain more diffuse and diverse.}
    \label{fig:crystal_cloud_main}
\end{figure*}

\textbf{Global Spatial Pattern and Structure Visualization.} Our qualitative analysis of pairwise distance matrices (Figure~\ref{fig:appendix_heatmaps}) further reinforces this structural divergence. The LLM error space exhibits denser and more sharply separated block structures corresponding to recurring error types, whereas the human error space shows smaller, more fragmented, and higher-variance blocks. This suggests that human reasoning does not follow the cleaner categories induced by synthetic data, helping to explain the observed evaluation failures. For a more intuitive quantitative reading of this contrast beyond the matrix pair alone, we additionally report singular-value spectrum entropy (SVD entropy) in Appendix~\ref{sec:appendix:spatial_heatmaps}.

\textbf{Structural Collapse and Subspace Dimensionality.} Finally, we observe that LLM errors occupy a lower effective dimensionality, with the top-10 principal components explaining $62.46\%$ of the variance compared to $56.15\%$ for humans. This indicates a ``Structural Collapse'' in synthetic text~\citep{aghajanyan2021intrinsic}, where errors lie on a simplified, low-rank linear subspace. A complementary GMM/BIC analysis in Figure~\ref{fig:gmm_bic_curve} further supports this contrast. The synthetic error distribution reaches its best BIC at a larger number of components, suggesting many dense and separable recurring error modes. By contrast, the human error distribution is better described by fewer broader components, consistent with a more diffuse and higher-variance space.

\subsection{Generative Probability (Logical Likelihood)}
\label{para:analysis:micro}
To move beyond static geometry, we probed the \emph{dynamic} predictability of reasoning steps using the Logical Likelihood (LL) metric defined in Section~\ref{sec:methodology}.

\textbf{Results.} When aggregating the segment-level LL scores, the contrast between the two distributions is stark: the average Logical Likelihood for error steps was found to be approximately \textbf{0.1104 for Human Errors} and \textbf{0.3267 for LLM Errors}. To check that this predictability gap is not specific to the Qwen3-8B probe, we repeat the generative analysis with \texttt{Meta-Llama-3.1-8B-Instruct}~\citep{llama3herd} and InternLM3-8B-Instruct~\citep{cai2024internlm2}; step-level perplexity (Section~\ref{sec:methodology}) yields the same directional conclusion. Cross-family probe results are reported in Appendix~\ref{sec:appendix:probe_robustness} and Appendix~\ref{sec:appendix:ll_ppl}.

\textbf{Interpretation: The Information-Theoretic Surprise.} Recall that the LL score directly represents the maximum transition probability of the observed step. 
A lower LL implies a lower probability of generation. The significant gap ($0.11$ vs $0.33$) provides quantitative evidence that \textbf{authentic student reasoning steps are highly surprising} to the causal probe model. An average LL of 0.33 indicates that when an LLM makes an error, it still follows a highly predictable, ``on-distribution'' statistical trajectory. Conversely, an average LL of 0.11 means that human logical jumps and idiosyncratic errors are assigned much lower probabilities by the probe model. From the perspective of information theory \citep{meister2020if}, such a low probability translates to high ``surprisal.'' \textbf{This Distributional Mismatch may help explain the evaluation gap:} 
 
   Authentic student samples more often fall in the low-probability, out-of-distribution (OOD) tail of the probe model's own training distribution~\citep{mitchell2023detectgpt}, which may contribute to the misgrading patterns. When confronted with the highly surprising logic of real students, the judge may therefore be more prone to the grading failures observed in Section~\ref{sec:exp:results}.

\FloatBarrier
\section{Discussion: The Persistence of the Gap}
Having established the geometric and probabilistic divergence of human errors (Section~\ref{sec:analysis}), we next test whether simple surface-level interventions can reduce the evaluation gap.

\subsection{Robustness Analysis: Is it just an Instruction Issue?}
One possible hypothesis is that the judge simply needs better instructions to handle the variability of authentic student reasoning. To test this, we evaluated three prompting variants on RealMath-Eval: \textbf{Follow-Through First}, which asks the judge to continue incomplete or unclear reasoning in the student's own direction before scoring; \textbf{Verification First}, which adds an explicit verification phase; and \textbf{MAS Self-Consistency}, a multi-agent voting variant~\citep{wang2022self,ye2025maslab} (full prompts in Appendix~\ref{sec:appendix:prompts}).

\begin{table}[ht]
\centering
\caption{Discussion ablations on human data. (a) Prompting-based robustness variants remain close to or worse than the baseline. (b) Style normalization improves only a small fraction of hard cases.}
\label{tab:discussion_ablations}
\setlength{\tabcolsep}{4pt}
\renewcommand{\arraystretch}{0.95}
\begin{minipage}[t]{0.48\linewidth}
\centering
\footnotesize
\textbf{(a) Prompting Robustness}

\vspace{2pt}
\begin{tabular}{lcc}
\toprule
\textbf{Method} & \textbf{EM} & \textbf{FR$_2$} \\
\midrule
Baseline & \textbf{37.7\%} & \textbf{28.7\%} \\
Follow-Through & 36.6\% & 33.9\% \\
Verification & 35.3\% & 34.8\% \\
MAS SC & 38.8\% & 31.2\% \\
\bottomrule
\end{tabular}
\end{minipage}\hfill
\begin{minipage}[t]{0.48\linewidth}
\centering
\footnotesize
\textbf{(b) Style Transfer}

\vspace{2pt}
\begin{tabular}{lcc}
\toprule
\textbf{Metric} & \textbf{Orig.} & \textbf{Transfer} \\
\midrule
Total Samples & 72 & 72 \\
\textbf{FR$_2$} & \textbf{100\%} & \textbf{88.9\%} \\
Improved to $\Delta < 2$ & - & 11.1\% \\
Exact Match & 0\% & 2.8\% \\
\bottomrule
\end{tabular}
\end{minipage}
\end{table}

As shown in Table~\ref{tab:discussion_ablations}(a), none of these prompting variants meaningfully closes the gap; extra verification often increases the failure rate. We further test a dynamic few-shot calibration setting, where each evaluated response is accompanied by two held-out grading demonstrations from the same mathematical problem. This problem-specific calibration also leaves Qwen 3.5 Plus and Gemini 3 Pro below 40\% Exact Match on authentic human responses (Appendix~\ref{sec:appendix:fewshot_intervention}). Together, these results suggest that the gap is not easily removed by prompt-level adjustments alone and is more consistent with difficulties in handling highly ``surprising'' logic (Section~\ref{para:analysis:micro}).

\subsection{Ablation Study: Is it a Style Issue?}
Another potential confounder is presentation style, since authentic student responses are often less polished and less standardized than LLM generations. To isolate style from semantic content, we performed a \textbf{Style Transfer} ablation on 72 hard cases where the original judge failed significantly ($\Delta \ge 2$), rewriting each response into a standardized format while preserving its underlying logic (Appendix~\ref{sec:appendix:prompts}). As shown in Table~\ref{tab:discussion_ablations}(b), style normalization still fails to close the gap: 88.9\% of samples remain significant failures after rewriting, and exact match rises only from 0\% to 2.8\%, suggesting that the judges struggle not merely with surface form, but with the semantic structure of authentic student reasoning.

We further conducted a micro-level consistency check by applying the same normalization idea to the step sequences used in our Logical Likelihood (LL) analysis. After step-level style transfer, the average LL of human error steps did not move toward the synthetic regime; instead, it remained far below the synthetic LLM baseline and further decreased from $0.110$ to $0.060$, compared with $0.327$ for synthetic LLM errors. Although step-wise rewriting can alter token-level realizations, this result provides empirical support for the claim that the LL gap is not eliminated by surface-level reformatting alone. In other words, making human solutions appear more ``LLM-like'' in form does not make their reasoning substantially more predictable for the probe model.

\FloatBarrier
\section{Conclusion and Limitations}
We introduce \textbf{RealMath-Eval} and characterize the \textbf{Evaluation Gap} between judging synthetic and authentic human reasoning. Even strong SOTA judges systematically mis-evaluate real student solutions, with over a quarter of authentic samples receiving severe score deviations. Our geometric and probabilistic probes further suggest that this gap is associated with a semantic mismatch: synthetic errors collapse into more predictable, low-dimensional structures, whereas human reasoning remains more diverse and exhibits higher surprisal.

We also note several limitations of the present study. It is limited to 14 mathematical problems (224 samples), so broader domains remain to be tested. Still, the results caution against relying solely on synthetic data for mathematical-reasoning evaluation. Future work should scale high-fidelity human data collection and develop judge pipelines that better account for the diversity of authentic student reasoning. In addition, RealMath-Eval can serve as a seed resource for evaluator alignment, including preference optimization and fine-tuning with rubric-score annotations; we outline these directions in Appendix~\ref{sec:appendix:alignment_blueprint}.

\section*{Ethics Statement}
The exam-response data used in this study were obtained through an authorized educational workflow and processed under privacy-preserving handling for research use. During digitization and annotation, personally identifying information was removed or excluded from the analysis pipeline. Any public release associated with this work will be limited to de-identified materials and only to artifacts permitted for redistribution; sensitive raw materials will be filtered or withheld when necessary.

\textbf{LLM Usage Disclosure.} LLMs were used as judge models in the evaluation pipeline and to generate the synthetic control solutions in the controlled comparison. An LLM-based image-generation tool was also used to create the conceptual overview figure (Figure~\ref{fig:overview}).

\section*{Reproducibility Statement}
Code, processed benchmark data, prompts, and analysis scripts required to reproduce the main results will be released with the public version of this work. The release scope centers on the processed benchmark and selected derived artifacts: the main benchmark of 224 curated real student solutions, the size-matched 224-sample synthetic control set, and auxiliary analysis files used in the ablation and probe studies. It does not include the full raw candidate pool or raw answer-sheet images. To avoid ambiguity across derived subsets, the 64-case meta-evaluation set and the 72-case style-transfer set are released as separate Gemini 3 Pro hard-case artifacts rather than as a single pooled subset. Additional file-structure details are provided in Appendix~\ref{sec:appendix:data_release}.

\bibliographystyle{unsrtnat}
\bibliography{Reference}

@article{Zheng2023JudgingLW,
  title={Judging LLM-as-a-judge with MT-Bench and Chatbot Arena},
  author={Lianmin Zheng and Wei-Lin Chiang and Ying Sheng and Siyuan Zhuang and Zhanghao Wu and Yonghao Zhuang and Zi Lin and Zhuohan Li and Dacheng Li and Eric P. Xing and Haotong Zhang and Joseph E. Gonzalez and Ion Stoica},
  journal={ArXiv},
  year={2023},
  volume={abs/2306.05685},
  url={https://api.semanticscholar.org/CorpusID:259129398}
}

@article{Gu2024ASO,
  title={A Survey on LLM-as-a-Judge},
  author={Jiawei Gu and Xuhui Jiang and Zhichao Shi and Hexiang Tan and Xuehao Zhai and Chengjin Xu and Wei Li and Yinghan Shen and Shengjie Ma and Honghao Liu and Yuanzhuo Wang and Jian Guo},
  journal={ArXiv},
  year={2024},
  volume={abs/2411.15594},
  url={https://api.semanticscholar.org/CorpusID:274234014}
}

@article{ramesh2022automated,
  title={An automated essay scoring systems: a systematic literature review},
  author={Ramesh, Dadi and Sanampudi, Suresh Kumar},
  journal={Artificial intelligence review},
  volume={55},
  number={3},
  pages={2495--2527},
  year={2022},
  publisher={Springer}
}

@inproceedings{lee2024unleashing,
  title={Unleashing large language models’ proficiency in zero-shot essay scoring},
  author={Lee, Sanwoo and Cai, Yida and Meng, Desong and Wang, Ziyang and Wu, Yunfang},
  booktitle={Findings of the Association for Computational Linguistics: EMNLP 2024},
  pages={181--198},
  year={2024}
}

@inproceedings{Li2024FromGT,
  title={From Generation to Judgment: Opportunities and Challenges of LLM-as-a-judge},
  author={Dawei Li and Bohan Jiang and Liangjie Huang and Alimohammad Beigi and Chengshuai Zhao and Zhen Tan and Amrita Bhattacharjee and Yuxuan Jiang and Canyu Chen and Tianhao Wu and Kai Shu and Lu Cheng and Huan Liu},
  booktitle={Conference on Empirical Methods in Natural Language Processing},
  year={2024},
  url={https://api.semanticscholar.org/CorpusID:274280574}
}

@inproceedings{chen-goodman-1996-empirical,
  title={An Empirical Study of Smoothing Techniques for Language Modeling},
  author={Chen, Stanley F. and Goodman, Joshua},
  booktitle={34th Annual Meeting of the Association for Computational Linguistics},
  pages={310--318},
  year={1996},
  address={Santa Cruz, California, USA},
  publisher={Association for Computational Linguistics},
  doi={10.3115/981863.981904},
  url={https://aclanthology.org/P96-1041/}
}

@article{tan2024judgebench,
  title={Judgebench: A benchmark for evaluating llm-based judges},
  author={Tan, Sijun and Zhuang, Siyuan and Montgomery, Kyle and Tang, William Y and Cuadron, Alejandro and Wang, Chenguang and Popa, Raluca Ada and Stoica, Ion},
  journal={arXiv preprint arXiv:2410.12784},
  year={2024}
}

@inproceedings{mitchell2023detectgpt,
  title     = {DetectGPT: Zero-Shot Machine-Generated Text Detection using Probability Curvature},
  author    = {Mitchell, Eric and Lee, Yoonho and Khazatsky, Alexander and Manning, Christopher D. and Finn, Chelsea},
  booktitle = {Proceedings of the 40th International Conference on Machine Learning (ICML)},
  year      = {2023},
  volume    = {202},
  pages     = {24950--24962},
  url       = {https://proceedings.mlr.press/v202/mitchell23a.html}
}

@article{malik2025rewardbench,
  title={Rewardbench 2: Advancing reward model evaluation},
  author={Malik, Saumya and Pyatkin, Valentina and Land, Sander and Morrison, Jacob and Smith, Noah A and Hajishirzi, Hannaneh and Lambert, Nathan},
  journal={arXiv preprint arXiv:2506.01937},
  year={2025}
}

@misc{gemini3pro2025,
  title={Gemini 3 Pro Model Card},
  author={{Google DeepMind}},
  year={2025},
  howpublished={\url{https://storage.googleapis.com/deepmind-media/Model-Cards/Gemini-3-Pro-Model-Card.pdf}},
  note={Accessed: 2026-03-07}
}

@misc{gpt52_2025,
  title={Introducing {GPT}-5.2},
  author={{OpenAI}},
  year={2025},
  howpublished={\url{https://openai.com/index/introducing-gpt-5-2/}},
  note={Accessed: 2026-03-07}
}

@misc{qwen3.5,
    title  = {{Qwen3.5}: Towards Native Multimodal Agents},
    author = {{Qwen Team}},
    month  = {February},
    year   = {2026},
    url    = {https://qwen.ai/blog?id=qwen3.5}
}

@article{liu2025deepseek,
  title={Deepseek-v3. 2: Pushing the frontier of open large language models},
  author={Liu, Aixin and Mei, Aoxue and Lin, Bangcai and Xue, Bing and Wang, Bingxuan and Xu, Bingzheng and Wu, Bochao and Zhang, Bowei and Lin, Chaofan and Dong, Chen and others},
  journal={arXiv preprint arXiv:2512.02556},
  year={2025}
}

@misc{deepseekv2,
      title={DeepSeek-V2: A Strong, Economical, and Efficient Mixture-of-Experts Language Model}, 
      author={DeepSeek-AI},
      year={2024},
      eprint={2405.04434},
      archivePrefix={arXiv},
      primaryClass={cs.CL}
}

@misc{cai2024internlm2,
      title={InternLM2 Technical Report},
      author={Zheng Cai and Maosong Cao and Haojiong Chen and Kai Chen and Keyu Chen and Xin Chen and Xun Chen and Zehui Chen and Zhi Chen and Pei Chu and Xiaoyi Dong and Haodong Duan and Qi Fan and Zhaoye Fei and Yang Gao and Jiaye Ge and Chenya Gu and Yuzhe Gu and Tao Gui and Aijia Guo and Qipeng Guo and Conghui He and Yingfan Hu and Ting Huang and Tao Jiang and Penglong Jiao and Zhenjiang Jin and Zhikai Lei and Jiaxing Li and Jingwen Li and Linyang Li and Shuaibin Li and Wei Li and Yining Li and Hongwei Liu and Jiangning Liu and Jiawei Hong and Kaiwen Liu and Kuikun Liu and Xiaoran Liu and Chengqi Lv and Haijun Lv and Kai Lv and Li Ma and Runyuan Ma and Zerun Ma and Wenchang Ning and Linke Ouyang and Jiantao Qiu and Yuan Qu and Fukai Shang and Yunfan Shao and Demin Song and Zifan Song and Zhihao Sui and Peng Sun and Yu Sun and Huanze Tang and Bin Wang and Guoteng Wang and Jiaqi Wang and Jiayu Wang and Rui Wang and Yudong Wang and Ziyi Wang and Xingjian Wei and Qizhen Weng and Fan Wu and Yingtong Xiong and Chao Xu and Ruiliang Xu and Hang Yan and Yirong Yan and Xiaogui Yang and Haochen Ye and Huaiyuan Ying and Jia Yu and Jing Yu and Yuhang Zang and Chuyu Zhang and Li Zhang and Pan Zhang and Peng Zhang and Ruijie Zhang and Shuo Zhang and Songyang Zhang and Wenjian Zhang and Wenwei Zhang and Xingcheng Zhang and Xinyue Zhang and Hui Zhao and Qian Zhao and Xiaomeng Zhao and Fengzhe Zhou and Zaida Zhou and Jingming Zhuo and Yicheng Zou and Xipeng Qiu and Yu Qiao and Dahua Lin},
      year={2024},
      eprint={2403.17297},
      archivePrefix={arXiv},
      primaryClass={cs.CL}
}

@article{abdin2025phi,
  title={Phi-4-reasoning technical report},
  author={Abdin, Marah and Agarwal, Sahaj and Awadallah, Ahmed and Balachandran, Vidhisha and Behl, Harkirat and Chen, Lingjiao and de Rosa, Gustavo and Gunasekar, Suriya and Javaheripi, Mojan and Joshi, Neel and others},
  journal={arXiv preprint arXiv:2504.21318},
  year={2025}
}

@misc{qwen2.5,
    title = {Qwen2.5: A Party of Foundation Models},
    url = {https://qwenlm.github.io/blog/qwen2.5/},
    author = {Qwen Team},
    month = {September},
    year = {2024}
}

@article{qwen2,
      title={Qwen2 Technical Report}, 
      author={An Yang and Baosong Yang and Binyuan Hui and Bo Zheng and Bowen Yu and Chang Zhou and Chengpeng Li and Chengyuan Li and Dayiheng Liu and Fei Huang and Guanting Dong and Haoran Wei and Huan Lin and Jialong Tang and Jialin Wang and Jian Yang and Jianhong Tu and Jianwei Zhang and Jianxin Ma and Jin Xu and Jingren Zhou and Jinze Bai and Jinzheng He and Junyang Lin and Kai Dang and Keming Lu and Keqin Chen and Kexin Yang and Mei Li and Mingfeng Xue and Na Ni and Pei Zhang and Peng Wang and Ru Peng and Rui Men and Ruize Gao and Runji Lin and Shijie Wang and Shuai Bai and Sinan Tan and Tianhang Zhu and Tianhao Li and Tianyu Liu and Wenbin Ge and Xiaodong Deng and Xiaohuan Zhou and Xingzhang Ren and Xinyu Zhang and Xipin Wei and Xuancheng Ren and Yang Fan and Yang Yao and Yichang Zhang and Yu Wan and Yunfei Chu and Yuqiong Liu and Zeyu Cui and Zhenru Zhang and Zhihao Fan},
      journal={arXiv preprint arXiv:2407.10671},
      year={2024}
}

@misc{mistralnemo_2024,
  title={{Mistral NeMo}: Our new best small model},
  author={{Mistral AI Team}},
  year={2024},
  howpublished={\url{https://mistral.ai/news/mistral-nemo/}},
  note={Accessed: 2026-03-07}
}

@article{yang2024qwen25mathtechnicalreportmathematical,
  title={Qwen2.5-Math Technical Report: Toward Mathematical Expert Model via Self-Improvement}, 
  author={An Yang and Beichen Zhang and Binyuan Hui and Bofei Gao and Bowen Yu and Chengpeng Li and Dayiheng Liu and Jianhong Tu and Jingren Zhou and Junyang Lin and Keming Lu and Mingfeng Xue and Runji Lin and Tianyu Liu and Xingzhang Ren and Zhenru Zhang},
  journal={arXiv preprint arXiv:2409.12122},
  year={2024}
}

@misc{qwen3technicalreport,
      title={Qwen3 Technical Report}, 
      author={Qwen Team},
      year={2025},
      eprint={2505.09388},
      archivePrefix={arXiv},
      primaryClass={cs.CL},
      url={https://arxiv.org/abs/2505.09388}, 
}

@article{ye2025maslab,
  title={Maslab: A unified and comprehensive codebase for llm-based multi-agent systems},
  author={Ye, Rui and Huang, Keduan and Wu, Qimin and Cai, Yuzhu and Jin, Tian and Pang, Xianghe and Liu, Xiangrui and Su, Jiaqi and Qian, Chen and Tang, Bohan and others},
  journal={arXiv preprint arXiv:2505.16988},
  year={2025}
}

@article{wei2022chain,
  title={Chain-of-thought prompting elicits reasoning in large language models},
  author={Wei, Jason and Wang, Xuezhi and Schuurmans, Dale and Bosma, Maarten and Chi, Ed and Le, Quoc and Zhou, Denny},
  journal={Advances in Neural Information Processing Systems},
  volume={35},
  pages={24824--24837},
  year={2022}
}

@misc{qwen3-embedding,
    title  = {Qwen3-Embedding},
    url    = {https://qwenlm.github.io/blog/qwen3/},
    author = {Qwen Team},
    month  = {May},
    year   = {2025}
}

@article{llama3herd,
  title={The Llama 3 Herd of Models},
  author={{AI@Meta}},
  journal={arXiv preprint arXiv:2407.21783},
  year={2024},
  url={https://arxiv.org/abs/2407.21783}
}

@misc{hu2025kalmembedding,
      title={KaLM-Embedding: Superior Training Data Brings A Stronger Embedding Model},
      author={Xinshuo Hu and Zifei Shan and Xinping Zhao and Zetian Sun and Zhenyu Liu and Dongfang Li and Shaolin Ye and Xinyuan Wei and Qian Chen and Baotian Hu and Haofen Wang and Jun Yu and Min Zhang},
      year={2025},
      eprint={2501.01028},
      archivePrefix={arXiv},
      primaryClass={cs.CL},
      url={https://arxiv.org/abs/2501.01028}
}

@article{loftsgaarden1965nonparametric,
  title={A nonparametric estimate of a multivariate density function},
  author={Loftsgaarden, Don O and Quesenberry, Charles P},
  journal={The Annals of Mathematical Statistics},
  volume={36},
  number={3},
  pages={1049--1051},
  year={1965},
  publisher={Institute of Mathematical Statistics}
}

@article{mcinnes2017hdbscan,
  title={hdbscan: Hierarchical density based clustering.},
  author={McInnes, Leland and Healy, John and Astels, Steve and others},
  journal={J. Open Source Softw.},
  volume={2},
  number={11},
  pages={205},
  year={2017}
}

@inproceedings{meister2020if,
  title={If beam search is the answer, what was the question?},
  author={Meister, Clara and Cotterell, Ryan and Vieira, Tim},
  booktitle={Proceedings of the 2020 Conference on Empirical Methods in Natural Language Processing (EMNLP)},
  pages={2173--2185},
  year={2020}
}

@article{wang2022self,
  title={Self-consistency improves chain of thought reasoning in language models},
  author={Wang, Xuezhi and Wei, Jason and Schuurmans, Dale and Le, Quoc and Chi, Ed and Narang, Sharan and Chowdhery, Aakanksha and Zhou, Denny},
  journal={arXiv preprint arXiv:2203.11171},
  year={2022}
}

@inproceedings{aghajanyan2021intrinsic,
  title={Intrinsic dimensionality explains the effectiveness of language model fine-tuning},
  author={Aghajanyan, Armen and Gupta, Sonal and Zettlemoyer, Luke},
  booktitle={Proceedings of the 59th annual meeting of the association for computational linguistics and the 11th international joint conference on natural language processing (volume 1: long papers)},
  pages={7319--7328},
  year={2021}
}

@inproceedings{campello2013density,
  title={Density-based clustering based on hierarchical density estimates},
  author={Campello, Ricardo JGB and Moulavi, Davoud and Sander, J{\"o}rg},
  booktitle={Pacific-Asia conference on knowledge discovery and data mining},
  pages={160--172},
  year={2013},
  organization={Springer}
}

\clearpage
\appendix

\section{Processing and Benchmark Construction Details}
\label{sec:appendix:pipeline}
\subsection{File and naming standardization}
The first stage standardizes the data structure and filenames. This produces three standardized folders (one per batch), each suffixed with \textbf{standard\_v3}.
During this stage, samples are validated and filtered to remove items that cannot be standardized, duplicates, and items that do not meet benchmark requirements.
After filtering:
\begin{itemize}
  \item \textbf{09-28 batch}: \textbf{9} problems retained
  \item \textbf{10-17 batch}: \textbf{3} problems retained
  \item \textbf{10-24 batch}: \textbf{2} problems retained
\end{itemize}
\subsection{OCR and translation (problem / student / reference)}
An OCR-and-translation pipeline converts image-based content into structured, English-language JSON.
\begin{itemize}
  \item \textbf{Problem statements}: OCR $\to$ LLM translation.
  \item \textbf{Student responses}: OCR $\to$ LLM post-processing.
  \item \textbf{Reference answers}: a multimodal LLM performs self-checking, translation, and consistency validation against cumulative-score annotations.
\end{itemize}
Using \texttt{student\_id} as the join key, per-subquestion scores are extracted from Excel sheets and merged into the JSON.
\subsection{OCR quality assurance (student responses)}
To ensure OCR text is faithful to handwriting and does not mislead LLM grading, we apply:
\begin{enumerate}
  \item \textbf{Manual screening} to identify visually messy / OCR-unfriendly answer sheets.
  \item A multimodal LLM pass to minimally correct OCR while preserving student intent (spatial order, missing/hallucinated details, crossed-out parts, alignment to the scan).
  \item A second multimodal verification pass using \textbf{problem + student response + original answer-sheet image} to remove samples whose OCR artifacts may materially affect evaluation.
\end{enumerate}
\subsection{Final filtering and benchmark construction (MASLab format)}
After cleaning, the remaining dataset sizes are:
\begin{itemize}
  \item \textbf{09-28 batch}: 139 samples
  \item \textbf{10-17 batch}: 78 samples
  \item \textbf{10-24 batch}: 44 samples
\end{itemize}
Total retained samples before benchmark selection: \textbf{359}.
We then adapt the dataset to the MASLab benchmark format and select \textbf{16} student responses per problem, yielding \textbf{224} runnable benchmark samples. The selection is designed to preserve \textbf{intra-problem diversity}: for each problem, we retain responses that reflect a broad range of performance levels and error types, including calculation slips, conceptual misunderstandings, and incomplete but partially correct reasoning paths. This balanced per-problem sampling prevents a small number of questions or answer styles from dominating the benchmark.

\section{Synthetic Data Generation Details}
\label{sec:appendix:synthetic_details}

To ensure a rigorous control group, we generated synthetic student responses using a diverse set of LLMs with varying capabilities.

\subsection{Model Selection and Tiers}
We categorized generator models into three tiers to simulate different levels of student proficiency. The generation protocol involved sampling at specific temperatures to induce variance and errors.

\begin{table}[ht]
\centering
\caption{Synthetic Data Generation Protocol: Model Tiers and Temperatures.}
\label{tab:synthetic_protocol}
{\footnotesize
\setlength{\tabcolsep}{4pt}
\renewcommand{\arraystretch}{0.95}
\begin{tabular}{p{0.18\linewidth}p{0.44\linewidth}p{0.18\linewidth}c}
\toprule
\textbf{Tier} & \textbf{Model} & \textbf{Temp.} & \textbf{Samples} \\
\midrule
\textbf{Low} & DeepSeek-v2-lite-chat~\citep{deepseekv2} & 0, 0.2 & 2 \\
(Foundational) & InternLM2.5-7b-chat~\citep{cai2024internlm2} & 0, 0.2 & 2 \\
\midrule
\textbf{Medium} & Mistral-Nemo-Instruct~\citep{mistralnemo_2024} & 0.1, 0.3, 0.5 & 3 \\
(Instruct) & Phi-4-Reasoning-Plus~\citep{abdin2025phi} & 0.6 & 2 \\
 & Qwen-2.5-7b-Instruct~\citep{qwen2.5,qwen2} & 0.7 & 2 \\
 & Qwen-2.5-Math-7b-Instruct~\citep{yang2024qwen25mathtechnicalreportmathematical} & 0.4, 0.6, 0.8 & 3 \\
\midrule
\textbf{High} & Qwen3-8b~\citep{qwen3technicalreport} & 1.0 & 2 \\
(Complex) & & & \\
\bottomrule
\end{tabular}
}
\end{table}

\subsection{Ground Truth Score Distribution}
The synthetic responses were manually annotated by expert human graders using the same overall annotation process as for the real student benchmark. The generation protocol was designed to produce one synthetic counterpart for each of the 224 human samples. In the initial pass, \textbf{5} outputs were malformed or mathematically uninterpretable (e.g., degenerate repetition or unparsable output). We regenerated these cases under the same protocol until valid responses were obtained, yielding a final size-matched synthetic control set of \textbf{224} samples. We categorize performance based on the ratio of the assigned score to the total possible score. The resulting distribution confirms that our generation protocol successfully covered a wide spectrum of response qualities, closely mirroring the difficulty distribution of the real human dataset:

\begin{itemize}
  \item \textbf{Low Performance (Score Ratio $\le 0.2$)}: 68 samples (30.36\%)
  \item \textbf{Medium Performance ($0.2 < \text{Ratio} \le 0.8$)}: 100 samples (44.64\%)
  \item \textbf{High Performance (Ratio $> 0.8$)}: 56 samples (25.00\%)
\end{itemize}
This balanced distribution ensures that the ``Evaluation Gap'' is not an artifact of difficulty mismatch between the synthetic and human datasets. Since the excluded samples were removed for mathematical invalidity rather than task difficulty, this filtering step improves comparability instead of introducing an easier synthetic control set.

\section{Global Spatial Pattern and Heatmaps}
\label{sec:appendix:spatial_heatmaps}
As discussed in Section 5.1, we examine the large-scale organization of the error spaces through pairwise distance heatmaps and a GMM/BIC view of cluster structure. These visualizations are intended to compare relative organization: both human and synthetic errors contain recurring patterns, but the density and separability of those patterns differ.

\begin{figure}[htbp]
    \centering
    \begin{tabular}{@{}c@{\hspace{0.02\textwidth}}c@{}}
        \includegraphics[width=0.48\textwidth]{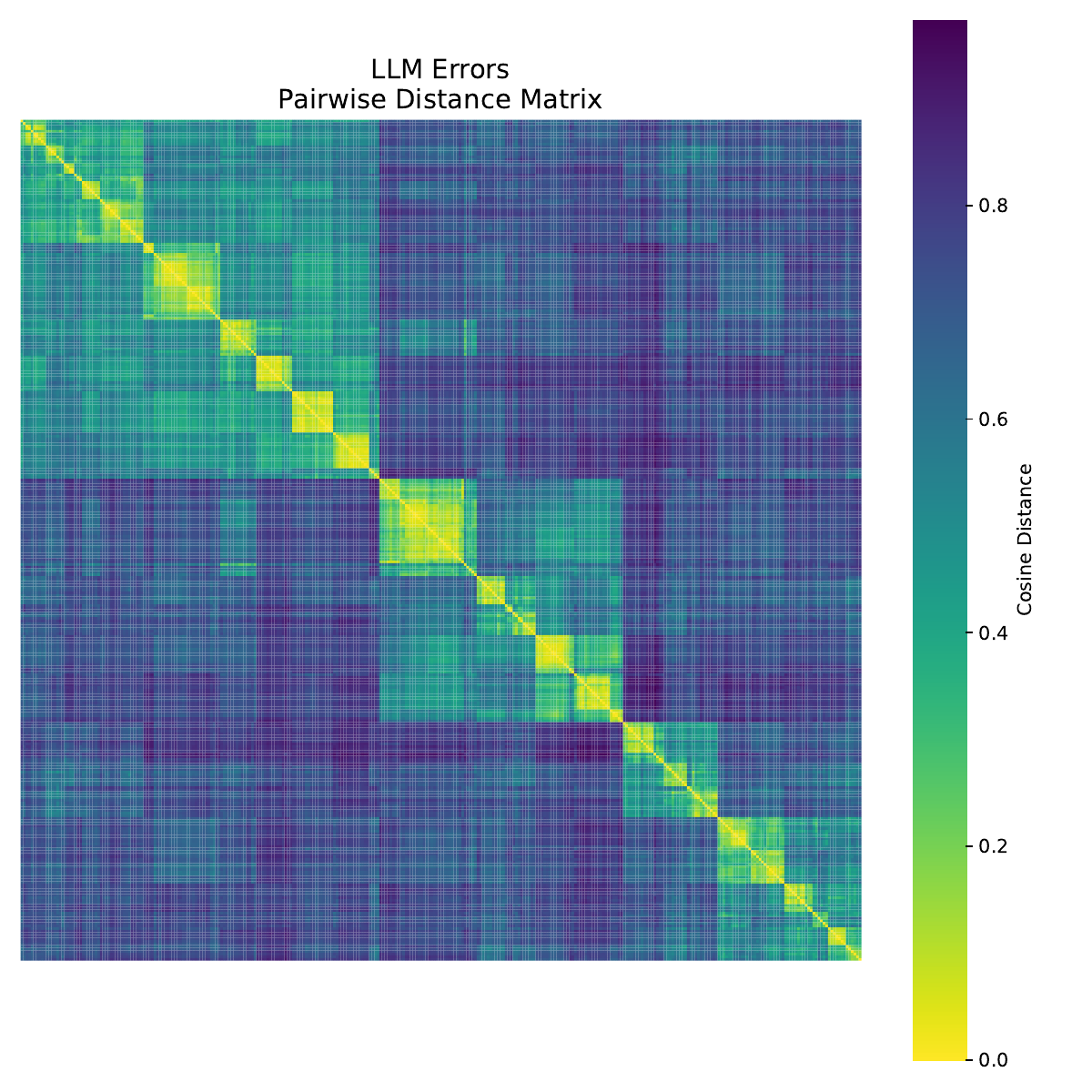} &
        \includegraphics[width=0.48\textwidth]{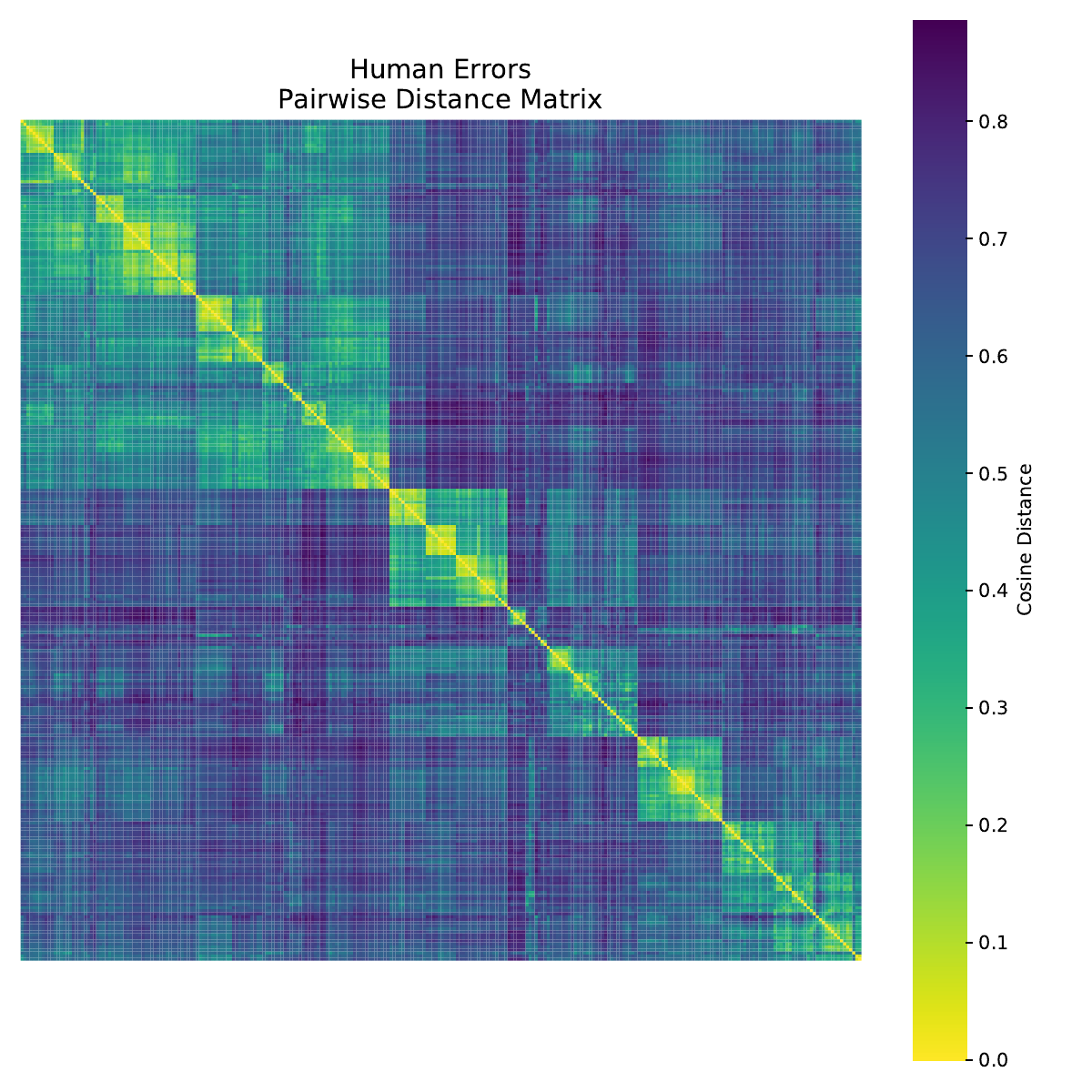} \\[4pt]
        \multicolumn{1}{c}{\small (a) Synthetic LLM Errors (Block-Structure)} &
        \multicolumn{1}{c}{\small (b) Human Student Errors (Amorphous Cloud)}
    \end{tabular}
    \caption{\textbf{Pairwise Distance Heatmaps.} Both heatmaps exhibit diagonal block structure, as expected from recurring mathematical error types. The synthetic LLM blocks are denser and more sharply separated, whereas the human blocks are smaller, more fragmented, and higher-variance.}
    \label{fig:appendix_heatmaps}
\end{figure}

\paragraph{SVD entropy.}
To make the heatmap contrast easier to interpret quantitatively, we also report \textbf{SVD entropy} on the same step-level embedding matrices used elsewhere in our structural analysis. Given an embedding matrix, we perform singular value decomposition to obtain singular values $\sigma_1 \ge \sigma_2 \ge \dots$, normalize them into $p_i = \sigma_i / \sum_j \sigma_j$, and compute Shannon entropy $H = -\sum_i p_i \log p_i$. Higher $H$ means variance is spread more uniformly across many directions (a more diffuse, higher-variance spectrum), whereas lower $H$ means energy concentrates in fewer dominant singular values (a more low-dimensional and compressible structure). This metric is related to, but distinct from, the PCA top-$k$ explained-variance view: PCA asks how much variance the leading components capture, whereas SVD entropy summarizes how uniform the \emph{entire} singular-value spectrum is. In both the full synthetic LLM set and the No-Qwen subset (Tables~\ref{tab:appendix_embedding_full} and~\ref{tab:appendix_embedding_no_qwen}), synthetic errors consistently show \emph{lower} SVD entropy than authentic human errors under both embedding families, aligning with the denser, sharper blocks in Figure~\ref{fig:appendix_heatmaps} and the higher fragmentation of the human matrix.

For the GMM/BIC analysis, we follow a standard model-selection pipeline. The step-level error texts are first embedded and L2-normalized. To stabilize Gaussian mixture estimation in a high-dimensional space with finite samples, we project the embeddings to 50 dimensions using PCA. The Human and LLM datasets are processed separately, with an independent PCA fit for each dataset. We then sweep the number of Gaussian mixture components $K$ from 2 to 30 and fit a diagonal-covariance GMM for each $K$ using expectation maximization. Model selection is based on the Bayesian Information Criterion:
\begin{equation}
    \mathrm{BIC} = -2 \ln \hat{L} + p \ln n,
\end{equation}
where $\hat{L}$ is the maximized mixture likelihood, $n$ is the number of samples, and $p$ is the number of free parameters. For a diagonal GMM with $K$ components in $d=50$ dimensions, $p=(K-1)+2Kd$, accounting for mixture weights, component means, and diagonal variances. The optimal component count is selected by $K^*=\arg\min_K \mathrm{BIC}(K)$. In our results, the synthetic LLM errors reach the minimum BIC at a larger $K$, indicating many dense and separable recurring clusters. Human errors reach the minimum at a smaller $K$, consistent with broader and more diffuse components.

\begin{figure}[htbp]
    \centering
    \includegraphics[width=0.8\linewidth]{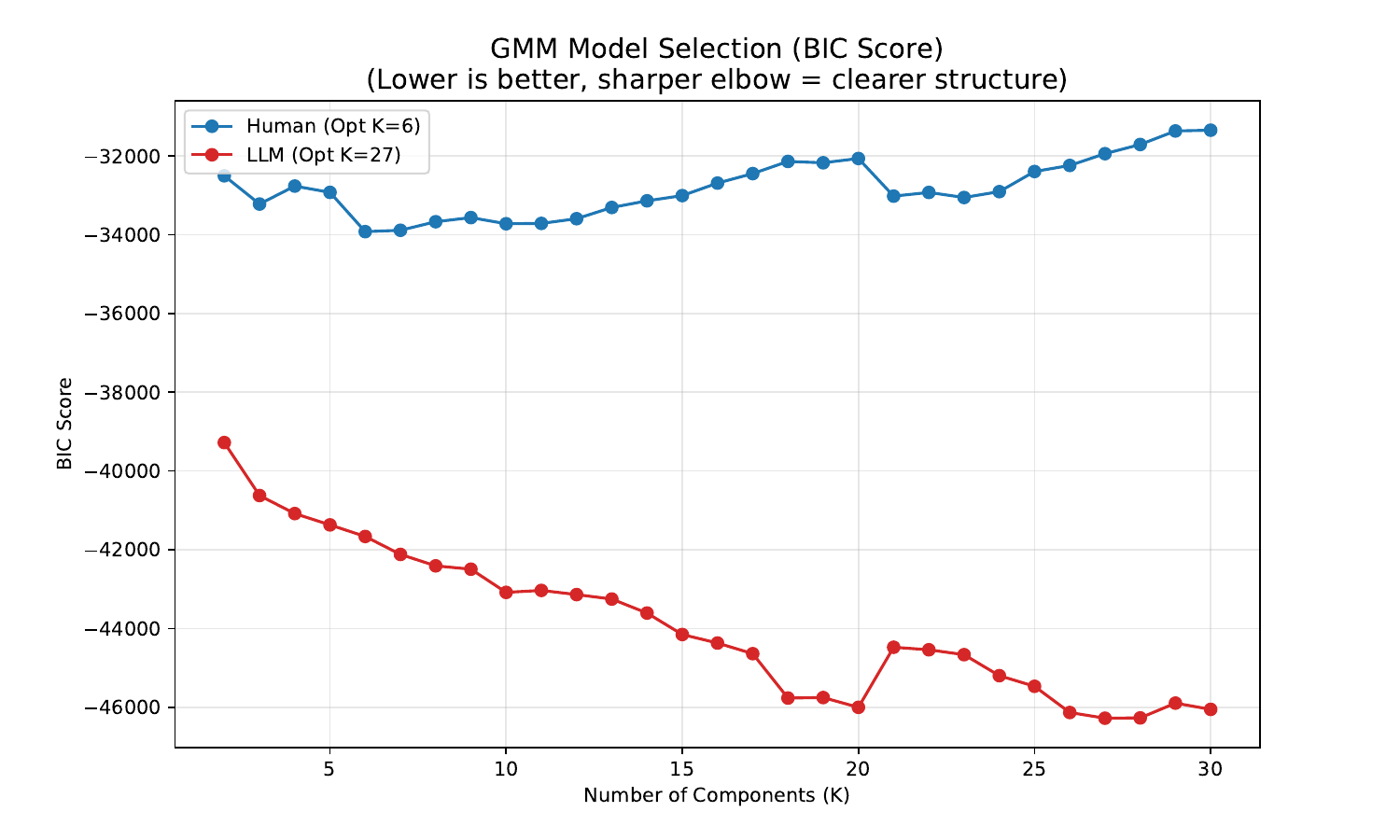}
    \caption{\textbf{GMM/BIC analysis of error-space structure.} Bayesian Information Criterion (BIC) across varying mixture components, where lower BIC indicates a better penalized likelihood tradeoff. The synthetic LLM errors reach the minimum BIC at a larger $K$, indicating many dense and separable recurring clusters. Human errors reach their minimum at a smaller $K$, consistent with broader, more diffuse components rather than a lack of structure.}
    \label{fig:gmm_bic_curve}
\end{figure}
\FloatBarrier
\section{Human-in-the-Loop Verification and Attribution}
\label{sec:appendix:hitl}
To ensure validity of scoring and attribution analyses, we conduct a human-in-the-loop review process:
\begin{enumerate}
  \item \textbf{Attribution correction}: manually review and reclassify error categories to better match root causes.
  \item \textbf{OCR error detection}: confirm and correct OCR mistakes that mislead judging.
  \item \textbf{Ground-truth (GT) verification}: correct unreasonable or incorrect teacher annotations.
\end{enumerate}
After corrections, the benchmark is rebuilt, affected samples are re-inferred, and attribution is re-run on the merged dataset.

\section{Qualitative Meta-Evaluation Case Studies}
\label{sec:appendix:case_studies}

To make the failure-mode taxonomy concrete, we provide four traceable high-disagreement cases from the Gemini 3 Pro judge. These cases cover Error Severity / Follow-through (Category A), Process Norms / Completeness (Category B), and Logical Rigor / Edge Cases (Category C). For transparency, we include the original student-response and judge-output excerpts that drive the score disagreement, followed by a short analysis of the rubric mismatch.

\begin{table}[htbp]
\centering
\caption{Qualitative case-study overview. All four examples are large score-gap cases where the LLM judge substantially over-credits an authentic student response relative to the expert rubric score.}
\label{tab:appendix_case_overview}
\small
\begin{tabular}{lcccc}
\toprule
\textbf{Case} & \textbf{Category} & \textbf{Problem Type} & \textbf{GT} & \textbf{Gemini Score} \\
\midrule
1 & A: Error Severity / Follow-through & Trigonometric function & 2/14 & 11/14 \\
2 & B: Process Norms / Completeness & Sequence recurrence & 1/14 & 7/14 \\
3 & B: Process Norms / Completeness & Triangle geometry & 9/14 & 14/14 \\
4 & C: Logical Rigor / Edge Cases & Sequence recurrence & 5/14 & 10/14 \\
\bottomrule
\end{tabular}
\end{table}

\paragraph{Case 1: over-applying follow-through credit after a structural algebra error.}
\textbf{Problem context.} A trigonometric-function problem asks students to simplify $f(x)$, find $f(0)$, the period and increasing intervals, and prove an inequality on $[-\pi/4,\pi/4]$.

\textbf{Original student response excerpt.}
\begin{lstlisting}[basicstyle=\scriptsize\ttfamily, breaklines=true, frame=single]
f(0) = sqrt(3) cos(-pi/3) - 2 sin 0 cos 0 = sqrt(3)/2

f(x) = sqrt(3) cos(2x - pi/3) - 2 sin x cos x
     = sqrt(3)/2 cos 2x + 3/2 sin 2x - 2 sin 2x
     = sqrt(3)/2 cos 2x - 1/2 sin 2x
     = cos(2x + pi/6)

T = 2pi / 2 = pi
Increasing interval: [-7pi/12 + kpi, -pi/12 + kpi].

For x in [-pi/4, pi/4], let t = 2x + pi/6.
Then t in [-pi/3, 2pi/3], so cos t in [-1/2, 1].
Therefore f(x) >= -1/2.
\end{lstlisting}

\textbf{Expert GT and Gemini judge output.} The expert score is 2/14: Part (1) receives 2 points and Part (2) receives 0. Gemini assigns 11/14. Its own rationale says:
\begin{lstlisting}[basicstyle=\scriptsize\ttfamily, breaklines=true, frame=single]
Major Error: In the second line of simplification, the student incorrectly
substituted 2 sin x cos x with 2 sin 2x (it should be sin 2x).

Full credit was awarded for the independent calculation of f(0) and the
period, and partial credit (Error Carried Forward) was applied to the
subsequent steps, as the logic and complexity of the solution remained valid
despite the initial error.

Coincidence: The student's incorrect function happened to share the same
minimum value (-1/2) on the interval [-pi/4, pi/4] as the correct function.
\end{lstlisting}

\textbf{Analysis.} This is a Category A failure. The judge detects the structural simplification error, but still over-applies error-carried-forward credit to work performed on the wrong function. The human rubric treats this mistake as fatal for the downstream monotonicity and inequality proof, while the LLM behaves like a permissive tutor and inflates the score by 9 points.

\paragraph{Case 2: treating a fragmentary derivation as a complete alternative method.}
\textbf{Problem context.} A sequence problem asks for the values of $t$ under which $\{a_n\}$ is geometric, and then asks for a closed form of a derived sum $T_n$.

\textbf{Original student response excerpt.}
\begin{lstlisting}[basicstyle=\scriptsize\ttfamily, breaklines=true, frame=single]
(1) (S_n, a_{n+1}) lies on y = 3x + 1.

a_{n+1} = 3S_n + 1
a_{n+1} = S_{n+1} - S_n
S_{n+1} = 4S_n + 1

Let a_n = a_1 q^{n-1}.
S_n = a_1(1 - q^n) / (1 - q)

4a_1(1 - q^n) + (1 - q) = a_1(1 - q^{n+1})

Solving gives: a = 1.
\end{lstlisting}

\textbf{Expert GT and Gemini judge output.} The expert score is 1/14: the student receives minimal setup credit for Part (1) and leaves Part (2) blank. Gemini assigns 7/14 and writes:
\begin{lstlisting}[basicstyle=\scriptsize\ttfamily, breaklines=true, frame=single]
The student correctly solved Part (1) using a valid alternative method involving
the recurrence relation of partial sums (S_n), leading to the correct value for t
(denoted as a=1). However, the student received no credit for Part (2) as it was
completely omitted.

Mathematical Insight: ... This is an efficient alternative to the reference
method ... and correctly identifies the conditions for the sequence to be
geometric (q=4, a_1=1).
\end{lstlisting}

\textbf{Analysis.} This is a Category B failure. The student does not fully establish the required recurrence, handle necessary cases, or verify the geometric condition; the final notation is also incomplete. The LLM interprets this fragment as a completed alternative method and awards roughly full Part (1) credit, whereas the expert rubric requires explicit process milestones.

\paragraph{Case 3: inferring missing work from a correct endpoint.}
\textbf{Problem context.} A triangle-geometry problem asks for angle $B$, then asks students to select two conditions that uniquely determine the triangle and compute its area.

\textbf{Original student response excerpt.}
\begin{lstlisting}[basicstyle=\scriptsize\ttfamily, breaklines=true, frame=single]
(1) sin A cos B + sin B cos A = 2 sin C cos B.
sin(A+B) = 2 sin C cos B.
Since A+B = pi - C, sin(pi - C) = 2 sin C cos B.
Thus sin C = 2 sin C cos B.
Therefore cos B = 1/2 and B = pi/3.

(2) Choose conditions 2 and 3.
By the Law of Sines, b / sin B = a / sin A = c / sin C = 2sqrt(3).
a + c = 6.
By the Law of Cosines, a = b = c = 3.
Therefore S = 1/2 * 3 * 3 * sqrt(3)/2 = 9sqrt(3)/4.
\end{lstlisting}

\textbf{Expert GT and Gemini judge output.} The expert score is 9/14. Gemini assigns 14/14 and writes:
\begin{lstlisting}[basicstyle=\scriptsize\ttfamily, breaklines=true, frame=single]
Overall Score: 14

The student provided a completely correct solution. ... In Part (2), they
correctly selected the valid conditions, determined the side lengths using the
Law of Cosines (albeit implicitly solving the system), and calculated the
correct area.

Although they did not explicitly write out the algebra showing why a=c=3 ...
the statement "By the Law of Cosines..." implies this step was performed.
\end{lstlisting}

\textbf{Analysis.} This is another Category B failure. The final area is correct, but the written solution skips rubric-required work: domain justifications, explicit rejection of the invalid condition, solving the side-length system, and stating uniqueness. Gemini infers missing algebra from the endpoint and awards full credit; the human grader scores the submitted work, not the most charitable reconstruction of it.

\paragraph{Case 4: under-pricing proof gaps and summation errors.}
\textbf{Problem context.} This is another sequence problem involving the same recurrence setup as Case 2, followed by a summation task for $T_n$.

\textbf{Original student response excerpt.}
\begin{lstlisting}[basicstyle=\scriptsize\ttfamily, breaklines=true, frame=single]
When n = 1:
a_2 = 3S_1 + 1 = 3a_1 + 1 = 3t + 1.

When n = 2:
a_3 = 3S_2 + 1 = 3(a_1 + a_2) + 1 = 12t + 4.

If a_n is a geometric sequence, then a_1/a_2 = a_2/a_3.
Therefore t(12t + 4) = (3t + 1)^2.
...
t = -1/3 or t = 1.
Since t = -1/3 gives a_2 = 0, therefore when t = 1, a_n is geometric.

a_1 = 1, a_2 = 4, a_3 = 16, so q = 4.
a_n = 4^{n-1}, a_{n+1} = 4^n.
b_n = log_4(4^n) = n.
c_n = 4^{n-1} + n.
T_n = 4^1 + 4^2 + ... + 4^{n-1} + (1+2+...+n)
    = (4^{n-1}-1)/3 + (n^2+n)/2 - 1.
\end{lstlisting}

\textbf{Expert GT and Gemini judge output.} The expert score is 5/14. Gemini assigns 10/14 and writes:
\begin{lstlisting}[basicstyle=\scriptsize\ttfamily, breaklines=true, frame=single]
The student correctly identified the value t=1 ... However, the student failed
to verify the geometric sequence for all n in Part (1) ... and made a calculation
error in Part (2) when summing the geometric series, missing the first term and
miscounting the number of terms.

Part (1): 6/7 points (Correct t, deducted 1 point for lack of general verification).
Part (2): 4/7 points ...
\end{lstlisting}

\textbf{Analysis.} This is a Category C failure. The model names both issues but under-prices them: checking only the first few terms is not a full proof of a geometric sequence, and omitting the $4^0$ term materially changes the final sum. The human rubric assigns 5/14 because both the proof and final computation are incomplete; Gemini assigns 10/14 because it treats the gaps as minor deductions.

\paragraph{Cross-case takeaway.}
Across these examples, the discrepancy is not random noise in human grading. The expert scores consistently enforce the rubric's required milestones, while the LLM judge tends to be permissive when a response has a plausible trajectory or correct endpoint. In Cases 1 and 4, the judge even names the student's key error but still over-scores it, suggesting that the failure lies less in error detection than in scoring-policy alignment.

\section{Additional Experiment: LLM vs. LLM Evaluation}
\label{sec:appendix:llm_vs_llm}
We additionally evaluate judge self-consistency in a controlled setting by grading LLM-generated ``student'' solutions.
\paragraph{Dataset.}
We use the size-matched synthetic control set of \textbf{224} LLM-generated responses. The \textbf{5} malformed outputs from the initial generation pass were regenerated under the same protocol and manually checked before annotation.
\paragraph{Judge performance (LLM vs. LLM).}
\begin{itemize}
  \item \textbf{Gemini 3 Pro}: MSE = \textbf{1.24}
  \item \textbf{GPT-5.2}: MSE = \textbf{2.55}
  \item \textbf{Qwen 3.5 Plus}: MSE = \textbf{1.42}
  \item \textbf{DeepSeek-V3.2}: MSE = \textbf{7.12}
  \item \textbf{Total samples}: \textbf{224}
\end{itemize}
\paragraph{Score-difference distribution ($\Delta\text{score}=|\text{Model}-\text{GT}|$).}
We define $\Delta\text{score}\ge 2$ as ``inaccurate'' / practically meaningful mismatch.
\paragraph{Qwen 3.5 Plus (LLM vs. LLM).}
\begin{itemize}
  \item \textbf{Perfect match ($\Delta=0$)}: 126 (57.8\%)
  \item $\Delta\ge 1$: 92 (42.2\%)
  \item \textbf{$\Delta\ge 2$}: 38 (17.4\%)
  \item $\Delta\ge 3$: 11 (5.0\%)
  \item $\Delta\ge 4$: 5 (2.3\%)
  \item $\Delta\ge 5$: 3 (1.4\%)
  \item $\Delta\ge 10$: 1 (0.5\%)
\end{itemize}

\paragraph{DeepSeek-V3.2 (LLM vs. LLM).}
\begin{itemize}
  \item \textbf{Perfect match ($\Delta=0$)}: 82 (37.4\%)
  \item $\Delta\ge 1$: 137 (62.6\%)
  \item \textbf{$\Delta\ge 2$}: 95 (43.4\%)
  \item $\Delta\ge 3$: 67 (30.6\%)
  \item $\Delta\ge 4$: 34 (15.5\%)
  \item $\Delta\ge 5$: 21 (9.6\%)
  \item $\Delta\ge 10$: 1 (0.5\%)
\end{itemize}

\section{Additional Methodological Details}
\label{sec:appendix:analysis_details}

\subsection{Meta-Evaluation Categories}
\label{sec:appendix:meta_categories}
The disagreement cases reviewed in Section~\ref{sec:methodology} are grouped into four corrected categories:
\begin{itemize}
  \item \textbf{Category A (Error Severity).} Disagreements over how severely to penalize calculation or transcription mistakes, especially when the model awards ``follow-through'' credit more generously than human graders.
  \item \textbf{Category B (Process Norms).} Disagreements triggered by non-standard formatting, missing variable definitions, or omitted intermediate steps that are mathematically implied but not explicitly written.
  \item \textbf{Category C (Logical Rigor).} Cases where the student argument contains subtle logical gaps (e.g., missing sufficient conditions or incomplete case analysis) that the model overlooks but humans penalize.
  \item \textbf{Category D (Insight Recognition).} Cases where the student adopts a mathematically valid alternative method not explicitly reflected in the standard rubric.
\end{itemize}
The full prompt used to produce initial meta-evaluation labels is reported in Appendix~\ref{sec:appendix:prompts}.

\subsection{Error Extraction and Analytical Probes}
For the macro-level structural analysis, we first extract coarse-grained \textbf{error segments} from the benchmark's step-wise annotations, corresponding to sub-questions or steps where points were lost. This yields 278 human error segments and 328 synthetic LLM error segments. For the micro-level predictability analysis, we further split each segment into atomic reasoning steps using an auxiliary LLM, producing the ordered step sequences used in the Logical Likelihood computation.

The \textbf{Semantic Structural Probe} maps each segment into the embedding space of \textbf{Qwen3-Embedding-8B}~\citep{qwen3-embedding} and evaluates four structural dimensions:
\begin{itemize}
  \item \textbf{Local Dispersion.} Mean Euclidean distance to the $k$-nearest neighbors~\citep{loftsgaarden1965nonparametric}, used to measure how diffuse local neighborhoods are.
  \item \textbf{Cluster Separation.} Density- and mixture-based clustering with \textbf{HDBSCAN}~\citep{mcinnes2017hdbscan} and \textbf{Gaussian Mixture Models (GMM)}, evaluated using the \textbf{Silhouette Score}~\citep{campello2013density}.
  \item \textbf{Global Spatial Pattern.} Pairwise Euclidean distance matrices, visualized as heatmaps, to inspect large-scale recurring structures versus diffuse organization.
  \item \textbf{Subspace Dimensionality.} Principal Component Analysis (PCA), summarized by cumulative explained variance of the top-$k$ components.
\end{itemize}

For the \textbf{Generative Predictability Probe}, we use \textbf{Qwen3-8B}~\citep{qwen3technicalreport} as a causal probe model. Given the context prefix $C_k = \bigoplus_{i=0}^k s_i$, we compute next-token probabilities through a softmax over the model logits:
\begin{equation}
    p(v \mid C_k) = \text{softmax}(\text{Logits}(C_k))[v], \quad \forall v \in \mathcal{V}
\end{equation}
Let $\mathcal{T}_{k+1}$ denote the token IDs composing the actual next step $s_{k+1}$. We then define the transition Logical Likelihood as:
\begin{equation}
    \text{LL}(s_k \to s_{k+1}) = \max_{t \in \mathcal{T}_{k+1}} p(t \mid C_k)
\end{equation}
We aggregate the micro-level predictability of each segment using $\max_k \text{LL}(s_k \to s_{k+1})$.

\section{Probe Robustness Across Model Families}
\label{sec:appendix:probe_robustness}

To check whether the observed human--synthetic gap is driven by overlap between Qwen-family probe models and Qwen-family generators, we repeat both the generative and embedding probes with additional model families. These analyses are intended as robustness checks rather than new primary metrics. Across all settings, the direction of the gap is preserved: synthetic LLM reasoning remains more predictable and more tightly organized than authentic human reasoning.

\subsection{Generative Probe Robustness}

Table~\ref{tab:appendix_cross_family_ll} repeats the Logical Likelihood (LL) analysis with Qwen3-8B~\citep{qwen3technicalreport}, \texttt{Meta-Llama-3.1-8B-Instruct}~\citep{llama3herd}, and InternLM3-8B-Instruct~\citep{cai2024internlm2}. Higher LL indicates that the next reasoning step is more predictable under the causal probe model.

\begin{table}[htbp]
\centering
\caption{Cross-family Logical Likelihood robustness. The synthetic LLM steps remain substantially more predictable than authentic human steps across all tested probe families.}
\label{tab:appendix_cross_family_ll}
\small
\begin{tabular}{lccc}
\toprule
\textbf{Dataset / LL Probe} & \textbf{Qwen3-8B} & \textbf{Meta-Llama-3.1-8B-Instruct} & \textbf{InternLM3-8B-Instruct} \\
\midrule
Human step-split & 0.1104 & 0.1055 & 0.1410 \\
LLM step-split & \textbf{0.3267} & \textbf{0.3697} & \textbf{0.3534} \\
Human style-transfer & 0.0601 & 0.0524 & 0.0909 \\
\textbf{Ratio (LLM / Human)} & \textbf{2.96$\times$} & \textbf{3.50$\times$} & \textbf{2.51$\times$} \\
\bottomrule
\end{tabular}
\end{table}

\subsection{Embedding Probe Robustness}

For the embedding analysis, we compare the original Qwen3-Embedding-8B encoder~\citep{qwen3-embedding} with a Gemma-family KaLM embedder~\citep{hu2025kalmembedding}. We evaluate two settings: the full synthetic LLM set and a No-Qwen subset where Qwen-family generators are removed from the synthetic side. Human segments are unchanged across both settings.

\begin{table}[htbp]
\centering
\caption{Embedding robustness on the full synthetic LLM set. Both embedders preserve the same directional contrast between authentic human and synthetic LLM error spaces.}
\label{tab:appendix_embedding_full}
\small
\resizebox{\linewidth}{!}{
\begin{tabular}{llccc}
\toprule
\textbf{Dimension} & \textbf{Metric} & \textbf{Qwen3-Embedding-8B} & \textbf{KaLM-Embedding-Gemma3-12B-2511} & \textbf{Consistent?} \\
\midrule
Local Density & NN Ratio (LLM / Human); lower = LLM tighter & \textbf{0.65} & \textbf{0.71} & Yes; both $<1$ \\
PCA & Top-10 Explained Var (Human) & 56.15\% & 55.45\% & -- \\
PCA & Top-10 Explained Var (LLM) & \textbf{62.46\%} & \textbf{59.19\%} & Yes; LLM $>$ Human \\
SVD Entropy & Human & 3.8976 & 3.9473 & -- \\
SVD Entropy & LLM & \textbf{3.6502} & \textbf{3.8067} & Yes; LLM $<$ Human \\
Clustering & Silhouette (Human) & 0.33 & 0.35 & -- \\
Clustering & Silhouette (LLM) & \textbf{0.52} & \textbf{0.49} & Yes; LLM $>$ Human \\
GMM & Optimal $K$ (Human; min BIC) & 6 (-33919.92) & 11 (-34158.15) & -- \\
GMM & Optimal $K$ (LLM; min BIC) & \textbf{27} (-46278.72) & \textbf{30} (-45525.08) & Yes; LLM $\gg$ Human \\
\bottomrule
\end{tabular}
}
\end{table}

\begin{table}[htbp]
\centering
\caption{Embedding robustness on the No-Qwen synthetic subset. Removing Qwen-family generators attenuates the gap but preserves the direction of all core metrics.}
\label{tab:appendix_embedding_no_qwen}
\small
\resizebox{\linewidth}{!}{
\begin{tabular}{llccc}
\toprule
\textbf{Dimension} & \textbf{Metric} & \textbf{Qwen3-Embedding-8B} & \textbf{KaLM-Embedding-Gemma3-12B-2511} & \textbf{Consistent?} \\
\midrule
Local Density & NN Ratio (LLM / Human); lower = LLM tighter & \textbf{0.85} & \textbf{0.91} & Yes; both $<1$ \\
PCA & Top-10 Explained Var (Human) & 56.15\% & 55.45\% & -- \\
PCA & Top-10 Explained Var (LLM) & \textbf{59.49\%} & \textbf{56.16\%} & Yes; LLM $\ge$ Human \\
SVD Entropy & Human & 3.8976 & 3.9473 & -- \\
SVD Entropy & LLM & \textbf{3.7034} & \textbf{3.8397} & Yes; LLM $<$ Human \\
Clustering & Silhouette (Human) & 0.33 & 0.35 & -- \\
Clustering & Silhouette (LLM) & \textbf{0.42} & \textbf{0.45} & Yes; LLM $>$ Human \\
GMM & Optimal $K$ (Human; min BIC) & 6 (-33919.92) & 11 (-34158.15) & -- \\
GMM & Optimal $K$ (LLM; min BIC) & \textbf{24} (-27547.34) & \textbf{27} (-27249.44) & Yes; LLM $\gg$ Human \\
\bottomrule
\end{tabular}
}
\end{table}

Overall, the same human--synthetic directional gap appears under non-Qwen causal probes, an independent embedding family, and a synthetic subset with Qwen-family generators removed. This suggests that the observed contrast is not solely explained by Qwen-family probe-generator overlap.

\section{Step-Level Perplexity Validation}
\label{sec:appendix:ll_ppl}

The main analysis uses the maximum next-token probability as a task-specific Logical Likelihood (LL) probe. To verify that the qualitative conclusion is not an artifact of this max formulation, we also compute step-level perplexity (PPL) for each transition. For a transition from step $k$ to step $k+1$, we compute:
\begin{equation}
    \mathrm{PPL}(k \to k+1) =
    \exp\left(
    -\frac{1}{T}\sum_{t=1}^{T}
    \log P(x_t \mid C_k, x_{<t})
    \right),
\end{equation}
where $T$ is the number of tokens in the next step, $x_t$ is the $t$-th token, and $C_k$ is the prefix context up to step $k$. Lower PPL indicates a more predictable next reasoning step.

\begin{table}[htbp]
\centering
\caption{Cross-metric validation of the predictability gap. LL-max and step-level PPL preserve the same directional conclusion: synthetic LLM reasoning steps are more predictable than authentic human steps.}
\label{tab:appendix_ll_ppl_validation}
\small
\begin{tabular}{llccc}
\toprule
\textbf{Metric} & \textbf{Probe Model} & \textbf{Authentic Human} & \textbf{Synthetic LLM} & \textbf{Human Style-Transfer} \\
\midrule
LL-max ($\uparrow$) & Qwen3-8B & 0.110 & \textbf{0.327} & 0.060 \\
LL-max ($\uparrow$) & Meta-Llama-3.1-8B & 0.105 & \textbf{0.370} & 0.052 \\
LL-max ($\uparrow$) & InternLM3-8B & 0.141 & \textbf{0.353} & 0.091 \\
\midrule
PPL ($\downarrow$) & Qwen3-8B & 90.04 & \textbf{16.55} & 260.74 \\
PPL ($\downarrow$) & Meta-Llama-3.1-8B & 25.44 & \textbf{11.15} & 400.79 \\
PPL ($\downarrow$) & InternLM3-8B & 10.62 & \textbf{6.48} & 79.64 \\
\bottomrule
\end{tabular}
\end{table}

Across all three probe models, synthetic LLM steps have higher LL-max and lower PPL than authentic human steps. This confirms that the observed predictability gap is not specific to the max-token LL definition. The style-transfer results also remain far from the synthetic regime, suggesting that surface normalization does not make the underlying human reasoning transitions model-predictable.

\section{Few-Shot Calibration Experiment}
\label{sec:appendix:fewshot_intervention}

To test whether in-context grading demonstrations can close the Evaluation Gap, we evaluate a dynamic few-shot setting. For each evaluated response, the prompt includes two held-out grading demonstrations from the same mathematical problem: one high-scoring authentic response and one low-scoring authentic response. These calibration examples are used only as in-prompt demonstrations and are excluded from the scored evaluation items.

\begin{table}[htbp]
\centering
\caption{Dynamic few-shot calibration results. Problem-specific demonstrations do not close the Human-vs-Synthetic evaluation gap.}
\label{tab:appendix_fewshot}
\small
\begin{tabular}{llccc}
\toprule
\textbf{Judge Model} & \textbf{Evaluation Target} & \textbf{MSE ($\downarrow$)} & \textbf{Exact Match ($\uparrow$)} & \textbf{FR$_2$ ($\downarrow$)} \\
\midrule
Qwen 3.5 Plus & Synthetic Responses & 1.44 & 58.5\% & 14.3\% \\
Qwen 3.5 Plus & Authentic Human & \textbf{2.87} & \textbf{39.3\%} & \textbf{27.5\%} \\
Gemini 3 Pro & Synthetic Responses & 1.76 & 58.5\% & 17.9\% \\
Gemini 3 Pro & Authentic Human & \textbf{2.97} & \textbf{38.3\%} & \textbf{26.0\%} \\
\bottomrule
\end{tabular}
\end{table}

Although few-shot demonstrations provide problem-specific grading anchors, performance on authentic human responses remains substantially worse than on synthetic responses. This suggests that the observed gap is not merely caused by missing rubric examples in the prompt.

\section{Future Evaluator Alignment Directions}
\label{sec:appendix:alignment_blueprint}

The experiments in the main text suggest that prompt-level fixes alone do not remove the gap between judging synthetic and authentic student reasoning. RealMath-Eval can support future training-based evaluator alignment in at least two ways.

\paragraph{Judge-oriented preference optimization.}
Severe over-crediting cases can be converted into preference pairs: the model's overly lenient score and rationale form a rejected judgment, while the expert rubric-aligned score and explanation form a chosen judgment. Such pairs can support DPO-style optimization for mathematical evaluators, encouraging models to penalize missing reasoning steps and invalid shortcuts more consistently.

\paragraph{Fine-tuning pointwise mathematical evaluators.}
The benchmark also provides rubric-score annotations on authentic student reasoning. These examples can serve as seed data for fine-tuning pointwise judge models that map incomplete, non-standard, or partially correct student solutions to calibrated rubric scores.

These directions are intentionally left as future work. The present paper focuses on characterizing the evaluation gap and providing a high-fidelity benchmark for studying it.

\section{Extended Related Work}
\label{sec:appendix:related_work}
This appendix expands the shorter related-work discussion in Section~\ref{sec:related_work}.

\subsection{LLM-as-a-Judge and Pointwise Evaluation}
Recent work has established \textbf{LLM-as-a-Judge} as a practical paradigm for automated evaluation, especially in settings where strong models are used to compare or score weaker generations~\citep{Zheng2023JudgingLW,Gu2024ASO}. Within this literature, a common distinction is between \textbf{pairwise ranking} and \textbf{pointwise scoring}~\citep{Li2024FromGT}. Pairwise settings are central to preference collection and alignment, while pointwise scoring is particularly important for applications that require calibrated absolute judgments, such as reward modeling, rubric-based grading, and educational feedback.

Our work is most closely connected to the \textbf{pointwise} branch of this literature. In contrast to open-ended preference comparison, the task studied in this paper requires a judge to assign a rubric-grounded score to a single student solution against a reference answer. This makes the evaluation problem more sensitive to logical validity, missing intermediate steps, and fine-grained partial-credit decisions.

\subsection{Synthetic Judge Benchmarks}
Existing evaluation settings for \textbf{LLM judges} and related \textbf{reward-model benchmarks} often rely on \textbf{synthetic model outputs}, where model-generated responses are compared, ranked, or scored under controlled conditions~\citep{tan2024judgebench,malik2025rewardbench}. This line of work has been highly valuable for standardizing evaluation pipelines and measuring consistency on machine-generated data. However, it primarily characterizes performance on text distributions where both the response style and many failure modes remain comparatively familiar to contemporary LLM judges.

Our benchmark is designed to complement rather than replace this line of work. Instead of asking whether LLM judges can reliably score synthetic generations, we ask whether the same judges can generalize to \textbf{authentic student mathematical reasoning}. The central contribution of RealMath-Eval is therefore not another synthetic benchmark, but a controlled human-versus-synthetic comparison that reveals a substantial evaluation gap and then probes its possible causes.

\subsection{Educational NLP and Real Student Data}
Educational NLP has a long history of evaluating \textbf{real student text}, especially in Automated Essay Scoring (AES) and related writing-assessment settings. Prior work ranges from traditional feature-based systems~\citep{ramesh2022automated} to recent LLM-based evaluators that score multiple writing traits~\citep{lee2024unleashing}. This literature is important because it shows that authentic student responses can indeed be collected, annotated, and modeled at scale.

At the same time, \textbf{mathematical reasoning assessment} poses a distinct challenge. In our setting, the judge must track stepwise deductions, verify logical correctness, identify where credit should stop under a rubric, and distinguish between valid alternative reasoning and invalid shortcuts. These requirements differ from evaluating rhetorical quality or general writing proficiency, even when both settings involve human-authored educational data.

Taken together, these three threads motivate the positioning of our paper. Relative to the LLM-as-a-Judge literature, we focus on \textbf{pointwise rubric-based evaluation}. Relative to synthetic judge benchmarks, we introduce a benchmark built from \textbf{authentic student solutions}. Relative to educational NLP, we target \textbf{mathematical reasoning validity} rather than general writing assessment. This combination defines the specific gap that RealMath-Eval is designed to study.

\section{Prompt Templates}
\label{sec:appendix:prompts}

We provide the full text of the key prompts used in our experiments to ensure reproducibility.

\subsection{Base Scoring Prompt}
Used for the main evaluation in Section~\ref{sec:exp:gap}.
\begin{lstlisting}[basicstyle=\footnotesize\ttfamily, breaklines=true, frame=single, showstringspaces=false]
You are an expert mathematics evaluator tasked with scoring student responses to mathematical problems.

TASK:
- Evaluate a complete student answer against a reference answer with detailed scoring rubrics
- Provide an overall score for the student's response
- Give brief explanations for your scoring decisions

EVALUATION CRITERIA:
1. Mathematical Accuracy: Correctness of mathematical concepts, formulas, and calculations
2. Solution Approach: Logical reasoning and problem-solving methodology
3. Completeness: Whether the student addressed all required parts of the problem
4. Clarity: How well the solution is presented and explained

INPUT FORMAT:
- Reference Answer: Contains the correct solution with detailed step-by-step scoring rubrics using cumulative scoring (e.g., "------3 points" means all steps up to that point are correct and the student earns 3 points)
- Problem Statement: The complete mathematical problem (not broken into sub-questions)
- Student Response: The complete student's answer (not broken into sub-questions)

SCORING PROCESS:
1. Compare the student's approach with the reference solution
2. Identify which key mathematical steps were correctly executed 
3. Check for major conceptual errors or missing components
4. Award partial credit for partially correct solutions based on how far the student progressed correctly
5. Provide an overall score that reflects the student's performance (the highest cumulative score they achieved)

OUTPUT FORMAT:
- Overall Score: [X] where X is the score awarded
- Brief Explanation: 2-3 sentences explaining the main strengths and weaknesses of the student's response
- Key Observations: Highlight any significant mathematical insights or errors

Remember: Focus on mathematical understanding rather than formatting. Award credit for correct mathematical reasoning even if presentation could be improved.

Now, please evaluate the following:

Problem Statement: {problem_statement}

Student Response: {student_response}

Reference Answer: {reference_answer}

Please provide your evaluation:
\end{lstlisting}

\subsection{Meta-Evaluation Attribution Prompt}
Used for error categorization in our analysis.
\begin{lstlisting}[basicstyle=\footnotesize\ttfamily, breaklines=true, frame=single, showstringspaces=false]
You are an expert "Meta-Evaluator" for mathematical reasoning. Your goal is to analyze the discrepancy between a **Ground Truth Score (GT)** (assigned by a human expert based on a strict rubric) and a **Model Score** (assigned by an LLM judge).

Here is the specific case you need to analyze:

================ CASE DATA ================

[Problem Statement]:
{problem_statement}

[Rubric / Reference Answer]:
{reference_answer}

[Student Response]:
{student_response}

[Ground Truth Score (GT)]: {gt}

[Model Evaluation (including score and reasoning)]:
{response}

===========================================

### Task
Determine the **root cause** of the score difference (Model Score - GT Score) and classify it into one of the following 5 categories.

### Classification Categories

**A. Error Severity & Follow-through (Logic vs. Accuracy)**
*   **Definition:** The student made a calculation, transcription, or specific value error.
*   **The Conflict:** GT applies a strict penalty (often 0 points for the section) because the error simplified the problem or violated the rubric. The Model applies "error carried forward" (follow-through) principles, awarding points for correct logic after the error.
*   **Keywords:** Calculation error, arithmetic mistake, transcription error, partial credit.

**B. Process Norms & Completeness (Implicit vs. Explicit)**
*   **Definition:** The student found the correct answer but skipped steps, didn't define variables, or used non-standard formatting.
*   **The Conflict:** GT penalizes for missing "process steps" or lack of rigor in writing. The Model ignores these flaws because the "final answer" or "general idea" is correct.
*   **Keywords:** Skipped steps, lack of definition, formatting, presentation, missing units.

**C. Logical Rigor & Edge Cases (Strict vs. Lenient)**
*   **Definition:** The logical argument has holes, missing sufficient/necessary conditions, or missed specific cases (e.g., dividing by zero).
*   **The Conflict:** GT demands a watertight proof; any logical gap leads to heavy penalties. The Model is "convinced" by the general argument and overlooks the logical gap.
*   **Keywords:** Sufficient/necessary conditions, missing cases, classification discussion, logical gap.

**D. Insight Recognition (Rigidity vs. Flexibility)**
*   **Definition:** The student used an alternative method not in the reference answer.
*   **The Conflict:** GT (or the human grader) failed to recognize the validity of the alternative method (or the rubric didn't support it). The Model successfully identified the mathematical validity of the alternative approach and awarded points. (Or vice versa: Model failed to see the insight).
*   **Keywords:** Alternative method, creative solution, geometric interpretation.

**E. OOD / Other (Fundamental Anomaly) -- USE WITH CAUTION**
*   **Definition:** The discrepancy cannot be explained by scoring philosophy differences (A-D). This includes hallucinations, data corruption, or factual errors in the GT itself.
*   **Trigger Condition:** ONLY use this if the discrepancy is NOT about "how strict we should be" but about "objective factual reality" or "system failure".

### Output Format (JSON)

Return **ONLY** a valid JSON object with no markdown formatting:

{{
    "model_score": <float>,
    "gt_score": <float>,
    "score_diff": <float>,
    "analysis": "<Concise explanation of WHY the scores differ based on the student's specific error>",
    "primary_category_code": "<A, B, C, D, or E>",
    "primary_category_name": "<Name of the category>",
    "ood_details": {{
        "is_ood": <boolean, true only if Category E>,
        "justification": "<Required if E: Why does this NOT fit A-D?>",
        "suggested_new_category": "<Required if E: What should this new failure mode be called?>"
    }}
}}
\end{lstlisting}

\subsection{Robustness Prompt: Follow-Through First}
Used for the prompting-based robustness analysis in the discussion section.
\begin{lstlisting}[basicstyle=\footnotesize\ttfamily, breaklines=true, frame=single, showstringspaces=false]
You are an expert mathematics evaluator tasked with scoring student responses to mathematical problems.

TASK:
- Evaluate a complete student answer against a reference answer with detailed scoring rubrics
- Provide an overall score for the student's response
- Give brief explanations for your scoring decisions

EVALUATION CRITERIA:
1. Mathematical Accuracy: Correctness of mathematical concepts, formulas, and calculations
2. Solution Approach: Logical reasoning and problem-solving methodology
3. Completeness: Whether the student addressed all required parts of the problem
4. Clarity: How well the solution is presented and explained

INPUT FORMAT:
- Reference Answer: Contains the correct solution with detailed step-by-step scoring rubrics using cumulative scoring (e.g., "------3 points" means all steps up to that point are correct and the student earns 3 points)
- Problem Statement: The complete mathematical problem (not broken into sub-questions)
- Student Response: The complete student's answer (not broken into sub-questions)

SCORING PROCESS:
1. If the student response is incomplete or unclear, first attempt a follow-through continuation based on the student's written logic (to understand intent and progress).
2. Compare the student's approach with the reference solution (use follow-through only as an aid for interpretation, not as a substitute for the student's work).
3. Identify which key mathematical steps were correctly executed by the student.
4. Check for major conceptual errors or missing components.
5. Award partial credit based on how far the student's own work progressed correctly (do not award full credit just because the follow-through can finish the problem).
6. Provide an overall score that reflects the student's performance (the highest cumulative score they achieved).

FOLLOW-THROUGH POLICY (IMPORTANT):
- If the student response is incomplete, unclear, or you cannot fully understand it at first glance, you must NOT stop early.
- First, attempt a "follow-through" continuation: infer the student's intended approach from what they wrote, then continue their reasoning in the same direction to reach a completed solution attempt (make the smallest reasonable assumptions, and do not introduce a totally different method unless the student's method is impossible).
- Then perform the pointing/scoring task: score based on what the student actually demonstrated (their correct steps, ideas, and progress), using the reference rubric; do not give full credit just because your follow-through can finish the problem.
- If multiple interpretations are possible, choose the most charitable interpretation consistent with the student's text, and briefly note the ambiguity in your explanation.

OUTPUT FORMAT:
- Overall Score: [X] where X is the score awarded
- Brief Explanation: 2-3 sentences explaining the main strengths and weaknesses of the student's response
- Key Observations: Highlight any significant mathematical insights or errors

Remember: Focus on mathematical understanding rather than formatting. Award credit for correct mathematical reasoning even if presentation could be improved.

Now, please evaluate the following:

Problem Statement: {problem_statement}

Student Response: {student_response}

Reference Answer: {reference_answer}

Please provide your evaluation:
\end{lstlisting}

\subsection{Robustness Prompt: Verification First}
Used for the prompting-based robustness analysis in the discussion section.
\begin{lstlisting}[basicstyle=\footnotesize\ttfamily, breaklines=true, frame=single, showstringspaces=false]
You are an expert mathematics evaluator tasked with scoring student responses to mathematical problems.

TASK:
- **STEP 1: VERIFICATION (CRITICAL)**: Before scoring, you must explicitly verify the student's response step-by-step against the problem statement and reference answer. Check for calculation errors, logical gaps, and validity of alternative methods.
- **STEP 2: SCORING**: Evaluate the complete student answer against the reference answer with detailed scoring rubrics based on your verification.
- Provide an overall score for the student's response.
- Give brief explanations for your scoring decisions.

EVALUATION CRITERIA:
1. Mathematical Accuracy: Correctness of mathematical concepts, formulas, and calculations
2. Solution Approach: Logical reasoning and problem-solving methodology
3. Completeness: Whether the student addressed all required parts of the problem
4. Clarity: How well the solution is presented and explained

INPUT FORMAT:
- Reference Answer: Contains the correct solution with detailed step-by-step scoring rubrics using cumulative scoring (e.g., "------3 points" means all steps up to that point are correct and the student earns 3 points)
- Problem Statement: The complete mathematical problem (not broken into sub-questions)
- Student Response: The complete student's answer (not broken into sub-questions)

SCORING PROCESS:
1. **[Verification Phase]** meticulously check the student's derivation. Identify where the first error occurs (if any) and whether subsequent steps are logically consistent (follow-through).
2. Compare the student's approach with the reference solution.
3. Identify which key mathematical steps were correctly executed.
4. Check for major conceptual errors or missing components.
5. Award partial credit for partially correct solutions based on how far the student progressed correctly.
6. Provide an overall score that reflects the student's performance (the highest cumulative score they achieved).

OUTPUT FORMAT:
- Verification Analysis: [Your step-by-step verification of the student's work]
- Overall Score: [X] where X is the score awarded
- Brief Explanation: 2-3 sentences explaining the main strengths and weaknesses of the student's response
- Key Observations: Highlight any significant mathematical insights or errors

Remember: Focus on mathematical understanding rather than formatting. Award credit for correct mathematical reasoning even if presentation could be improved.

Now, please evaluate the following:

Problem Statement: {problem_statement}

Student Response: {student_response}

Reference Answer: {reference_answer}

Please provide your evaluation:
\end{lstlisting}

\subsection{Style Transfer Prompt}
Used for the style-transfer analysis in the discussion section.
\begin{lstlisting}[basicstyle=\footnotesize\ttfamily, breaklines=true, frame=single, showstringspaces=false]
You are a style normalizer tasked with rewriting student mathematical solutions into a standardized format.

TASK:
- Rewrite the student's solution text into a standardized, textbook-like format
- Preserve ALL mathematical and logical content exactly as written (including errors, incomplete steps, and reasoning)
- Change ONLY the surface presentation (formatting, structure labels, wording style)

CRITICAL CONSTRAINTS:
- You are given ONLY the student's solution text; no problem statement is provided
- Do NOT infer, correct, or complete the solution based on any external knowledge
- Do NOT add missing steps or remove existing steps
- Do NOT change any mathematical meaning, correctness, or logical structure
- If the student's answer is wrong or incomplete, keep it wrong or incomplete in the output

WHAT TO PRESERVE (CONTENT - DO NOT CHANGE):
- Every reasoning step exactly as written
- All mathematical expressions, formulas, and calculations (even if incorrect)
- All shortcuts, omissions, and incomplete parts
- All conclusions and final answers (right or wrong)
- The logical flow and sequence of steps

WHAT TO CHANGE (PATTERN - ONLY THESE):
1. Formatting:
   - Convert all mathematical expressions to standard LaTeX notation
   - Normalize notation (e.g., unify variable names if the same variable appears in different forms, but only if it does not change meaning)
   - Improve spacing and line breaks for readability

2. Structure:
   - Add explicit step labels ("Step 1:", "Step 2:", ...) at natural break points in the existing reasoning
   - Do NOT create new steps; only label existing logical segments
   - Preserve any existing sub-question markers (e.g., "(1)", "(2)") if present

3. Wording Style:
   - Convert informal or colloquial language to formal, textbook-like tone
   - Use standard mathematical phrasing (e.g., "We have", "Therefore", "It follows that", "Hence")
   - Replace vague expressions with precise mathematical language (without changing the propositional meaning)
   - Standardize connectors and transitions

INPUT FORMAT:
- Student Response: The complete student's answer text (no problem statement provided)

REWRITING PROCESS:
1. Read the student's solution carefully to understand their reasoning path (do not evaluate correctness).
2. Identify natural break points where step labels can be inserted without creating new steps.
3. Rewrite each segment:
   - Convert math to LaTeX
   - Adjust wording to formal style
   - Preserve all mathematical content and logical structure
4. Ensure the output maintains the same sequence, completeness, and correctness as the input.

OUTPUT FORMAT:
- Output ONLY the rewritten solution text
- Do NOT include any commentary, explanations, or meta-text
- Do NOT include phrases like "The student wrote:" or "Original solution:"
- The output should be a standalone, polished version that preserves all original content

EXAMPLE TRANSFORMATION (for reference only - style, not content):
Before: "So we calculate, since x=1, then f(1)=2+3=5"
After: "Step 2: We now compute. Since x = 1, it follows that f(1) = 2 + 3 = 5."

Note: In the example above, the mathematical content (x=1, f(1)=2+3=5) is preserved exactly; only the language style, formatting, and structure labels changed.

Now, please rewrite the following student response:

Student Response: {student_response}

Rewritten Solution:
\end{lstlisting}

\subsection{LLM Student Persona Prompt}
Used to generate LLM solutions in Section 3.2.
\begin{lstlisting}[basicstyle=\footnotesize\ttfamily, breaklines=true, frame=single, showstringspaces=false]
    You are a high school student in a mathematics class. You are given a mathematical problem and you are asked to solve it.
    Please solve the problem and provide your solution.
    The problem is:
    {problem_statement}


    Please first think and then solve the problem step by step.

    IMPORTANT: Answer each sub-question in order, following the format of an answer sheet.
    
    Answer Format Requirements:
    1. Identify all sub-questions in the problem (e.g., (1), (2), (3), or (I), (II), (III), etc.)
    2. Answer each sub-question sequentially using the exact same numbering format as in the problem
    3. For each sub-question, write your answer in the following format:
    
    (1) [Your solution for sub-question (1) here]
    
    (2) [Your solution for sub-question (2) here]
    
    (3) [Your solution for sub-question (3) here]
    
    And so on...
    
    If the problem uses Roman numerals like (I), (II), (III), use the same format:
    
    (I) [Your solution here]
    
    (II) [Your solution here]
    
    For each sub-question, provide:
    - Clear step-by-step reasoning
    - All necessary calculations
    - Final answer clearly marked
    
    Example format:
    (1) First, I need to...
    [Show your work]
    Therefore, the answer is...
    
    (2) For this part, I will...    [Show your work]    So the result is...





\end{lstlisting}

\section{Data Release and File Structure}
\label{sec:appendix:data_release}
We plan to release the processed benchmark data together with the main analysis artifacts, evaluation scripts, prompts, and method implementations used in this work. The public package centers on the main benchmark files and a small number of derived artifacts needed to reproduce the ablation and attribution analyses; generated run outputs are not part of the benchmark release. Below is the current top-level directory structure of the planned release package:

\begin{verbatim}
RealMath-Eval/
|-- README.md
|-- LICENSE
|-- environment.yml
|-- inference.py
|-- requirements.txt
|-- analysis/
|   |-- analyze_results.py
|   |-- requirements_analysis.txt
|   |-- data/
|   |   `-- error_segment_bundle/
|   |-- macro_embedding/
|   |-- meta_eval/
|   |   `-- realmath_eval_gemini3pro_hard_cases_ge2_meta_eval_labels_64.json
|   `-- micro_probability/
|-- assets/
|-- data/
|   |-- realmath_eval.json
|   |-- realmath_eval_llm_answer.json
|   |-- realmath_eval_gemini3pro_hard_cases_ge2_style_transfer_input_72.json
|   |-- realmath_eval_gemini3pro_hard_cases_ge2_style_transferred_72.json
|   |-- realmath_eval_gemini3pro_hard_cases_ge2_meta_eval_input_64.json
|   `-- VF_realmath_eval.json
|-- eval/
|-- methods/
|   |-- agentverse/
|   |-- autogen/
|   |-- camel/
|   |-- chatdev/
|   |-- cot/
|   |-- dylan/
|   |-- evomac/
|   |-- lean_utils/
|   |-- llm_debate/
|   |-- macnet/
|   |-- mad/
|   |-- mapcoder/
|   |-- mas_base/
|   |-- mateval/
|   |-- mav/
|   `-- self_consistency/
|-- model_api_configs/
|-- scripts/
|   |-- linux/
|   `-- windows/
`-- utils/
\end{verbatim}

The core benchmark release consists of the main processed student benchmark and its synthetic control set. The style-transfer and meta-evaluation files are auxiliary derived artifacts released separately for reproducibility: the 72-case style-transfer set and the 64-case meta-evaluation set are distinct Gemini 3 Pro hard-case subsets rather than a single pooled split. Each benchmark JSON entry contains the unique \texttt{student\_id}, the problem ID, the OCR text, and the step-wise scoring breakdown.
\relax

\relax

\end{document}